\documentclass[Afour,sageh,times]{sagej}
\usepackage{tabularx}
\usepackage[subrefformat=parens,labelformat=parens]{subfig}
\usepackage{amssymb}
\usepackage{amsmath}
\usepackage{bm}      
\usepackage{subfig}
\usepackage{times}
\usepackage{graphicx}
\usepackage{flushend}
\usepackage{algorithm}
\usepackage{algpseudocode}
 
\usepackage[bookmarks=true]{hyperref}
\DeclareMathOperator*{\argmax}{arg\,max}

\algdef{S}[FOR]{ForEach}[1]{\algorithmicforeach\ #1\ \algorithmicdo}
\algnewcommand\algorithmicinput{\textbf{Input:}}
\algnewcommand\INPUT{\item[\algorithmicinput]}
\algnewcommand\algorithmicoutput{\textbf{Output:}}
\algnewcommand\OUTPUT{\item[\algorithmicoutput]}

\usepackage{xcolor}

\hypersetup{
    colorlinks,
    linkcolor={red!50!black},
    citecolor={blue!50!black},
    urlcolor={blue!80!black}
}

\graphicspath{{./fig/}{./fig/new-exp/}{./fig/appendix/}}
\DeclareGraphicsExtensions{.pdf,.png,.jpg}

\newcommand\BibTeX{{\rmfamily B\kern-.05em \textsc{i\kern-.025em b}\kern-.08em
T\kern-.1667em\lower.7ex\hbox{E}\kern-.125emX}}

\def\volumeyear{2023}

\begin{document}

\runninghead{}

\title{Boundary-Aware Value Function Generation for Safe Stochastic Motion Planning
\vspace{10pt}
}

\author{
Junhong Xu\affilnum{1}, Kai Yin\affilnum{2}, Jason M. Gregory\affilnum{3}, Kris Hauser\affilnum{4}, Lantao Liu\affilnum{1}
}

\affiliation{\affilnum{1}Luddy School of Informatics, Computing, and Engineering, Indiana University-Bloomington.\\
\affilnum{2}Expedia Group. \\
\affilnum{3}U.S. Army Research Laboratory.\\
\affilnum{4}Department of Computer Science, University of Illinois at Urbana-Champaign
}

\corrauth{Lantao Liu, Luddy School of Informatics, Computing, and Engineering, Indiana University, Bloomington, IN 47408.
}

\email{lantao@iu.edu}

\begin{abstract}
Navigation safety is critical for many autonomous systems such as self-driving vehicles in an urban environment. It requires an explicit consideration of boundary constraints that describe the borders of any infeasible, non-navigable, or unsafe regions. We propose a principled boundary-aware safe stochastic planning framework with promising results.  Our method generates a value function that can strictly distinguish the state values between free (safe) and non-navigable (boundary) spaces in the continuous state, naturally leading to a safe boundary-aware policy.  At the core of our solution lies a seamless integration of finite elements and kernel-based functions,  where the finite elements allow us to characterize safety-critical states' borders accurately, and the kernel-based function speeds up computation for the non-safety-critical states.  The proposed method was evaluated through extensive simulations and demonstrated safe navigation behaviors in mobile navigation tasks. Additionally, we demonstrate that our approach can maneuver safely and efficiently in cluttered real-world environments using a ground vehicle with strong external disturbances, such as navigating on a slippery floor and against external human intervention. 
\end{abstract}


\maketitle

\section{Introduction}
Successful deployments of mobile navigation tasks, such as last-mile delivery and autonomous driving, require the robot to achieve safe 
and efficient motions in complex and cluttered environments.
This task is challenging 
due to a mismatch between the actual vehicle dynamics and the approximated model utilized for motion planning.
The uncertainty during the execution also requires planning for a feedback policy rather than a single trajectory.

As an important foundation of many decision-making under uncertainty and stochastic planning problems, the Markov Decision Process (MDP) 
has been utilized to address the robot motion planning and control problem in complex environments under disturbances~\citep{puterman2014markov}. 
{Extending MDPs to continuous states for real-world robotics applications necessitates the approximation of the value function and/or policy.}
A widely adopted approach is to leverage continuous function approximations such as neural networks or kernel functions~\citep{ormoneit2002kernel, fan2020theoretical, geist2013algorithmic} to fit the continuous states in the discrete MDP framework. 
However, these approximations both worsen the policy optimality and exacerbate the collision violation because the 
true optimal value function can exhibit sharp gradients towards non-navigable boundaries or even discontinuities at constraint borders, but the approximations typically cannot effectively characterize (actually blur) the {values at the boundaries between the navigable (safe) regions and those non-navigable (unsafe) spaces. 

In this work, we propose a principled value function representation framework for continuous MDPs 
that guarantee 
boundary conditions. 
We model the safety-constrained decision problem with a variant of continuous-state MDP -- diffusion MDP (DMDP) -- whose optimal solution is characterized by a partial differential equation (PDE)~\citep{Xu-RSS-20}. 
This allows us to embed the environmental geometries and task constraints into boundary conditions coupled with the PDE.  
Since the boundary conditions accompany the PDE, we can naturally embed the safety and the goal constraints into the value function.

We propose to utilize the Finite Element Method (FEM) to approximate the value function and solve the PDE numerically. 
FEM discretizes the entire state space by subdividing it into a ``mesh" of discrete elements. It is used to approximate the value function by creating local basis functions within each element. 
An important feature of these basis functions is their compact support, spanning only a few elements and thus having limited overlap with other basis functions. This characteristic enables the value function to respond sensitively to local changes in state.
We show that, 
even with simple basis functions, FEM excels in accurately characterizing safety-critical boundaries. 
This provides an extremely high-quality approximation for satisfying the safety-critical boundary conditions which existing value function approximation methods cannot achieve.  

Moreover, we develop a novel mechanism that blends the benefits of ``mesh-based" and ``meshless" based methods within the FEM framework, leading to a new tensor-product basis function. 
The meshless part in the basis function leverages {\em kernel} representation which alleviates the burden of numerical integrals in the FEM framework.
This mechanism is actually motivated by the observation that boundary conditions are imposed only on some, not all, state dimensions in many robotics applications. 
For instance, geometric obstacle constraints are typically imposed on the robot's position rather than its velocity profiles.
Consequently, when safety and boundary avoidance are a priority for all dimensions, a pure mesh-based method can be utilized; when specific dimensions do not require strict safety satisfaction or goal constraints, the meshless kernel-based method can be employed {combined with finite elements} to speed up the computation. 
 
In a range of diverse simulated environments featuring varying obstacle ratios, we conducted experiments to demonstrate our proposed method in maintaining safe and goal-reaching robotic motion when compared to other approaches.
We also tested and validated the proposed method using an autonomous ground vehicle in real-world environments with varying obstacle configurations and strong external disturbances.  
The results show that our method can perform safe and fast maneuvers in {\em all} tested real-world trials.


\section{Related Work}
Our research centers on the computation of a boundary-aware value function to establish a safe feedback policy for robot planning under motion uncertainty. In this section, we survey related literature on motion planning and control for such uncertain systems, emphasizing approaches that ensure safety and effectively accomplish goal-oriented objectives.

\subsection{Deterministic and Stochastic Motion Planning}\label{sec:related-work-motion-planning}
Deterministic motion planning typically employs deterministic robot motion models to compute open-loop paths~\citep{gammell4asymptotically}. 
Due to modeling errors between the model used for planning and the real world, the robot usually cannot directly execute the computed path.
Thus, it becomes essential to refine it into a kinodynamically feasible trajectory~\citep{webb2012kinodynamic} for a low-level controller to track. 
One common approach to generating such feasible motion trajectories involves trajectory optimization.
We can generally categorize existing trajectory optimization techniques into hard-constrained and soft-constrained methods. 
Hard-constrained methods, e.g.,~\citet{mellinger2011minimum}, generate various geometric volumes such as cubes \citep{chen2015real, gao2018online}, spheres \citep{gao2016online, gao2019flying}, or polyhedra \citep{liu2017planning, deits2015efficient, richter2016polynomial}, within the safe movement region, and compute the smooth trajectories within the volumes.
As long as these geometric volumes do not overlap with obstacles, this approach can force the robots to keep a distance from obstacles to avoid a collision, though sometimes it produces relatively conservative trajectories. 
On the other hand, instead of formulating safety into constraints, soft constraint methods consider safety along with other task objectives, such as trajectory smoothness, in the cost function.
For example, distance gradients are usually utilized during optimization to ensure that the trajectory keeps a safe distance from obstacle boundaries~\citep{zhou2020robust, oleynikova2016continuous, oleynikova2016signed, zhou2019robust,ratliff2009chomp}.
While soft constraint methods may not provide an absolute guarantee of safety, they do offer a more adaptable optimization process. This flexibility allows for a degree of constraint violation, proving advantageous in securing a feasible solution where rigid adherence to hard constraints might render the problem infeasible.

The trajectory optimization techniques outlined earlier mainly rely on a deterministic model to generate dynamically feasible motion trajectories. In contrast, stochastic planning presents a more robust solution to motion uncertainties.
It employs a stochastic motion model that accounts for uncertainties arising from modeling inaccuracies or unforeseen environmental changes.
Early work within the robotics community that merged stochastic planning and control predominantly relied on the sampling-based paradigm~\citep{lavalle2006planning, karaman2011sampling}. 
For instance, \citet{tedrake2010lqr} computed local feedback controllers along expanding new branches in the rapidly exploring random tree (RRT) algorithm to stabilize the robot's movement between two tree nodes. 
A comparable strategy was employed by \citet{agha2014firm} but with a different planner and controller. Their method considered both sensing and motion uncertainty, utilizing the linear-quadratic Gaussian (LQG) controller~\citep{kalman1960contributions} to stabilize motion and sensing uncertainties along the edges created by the probabilistic roadmap (PRM) algorithm. 
An alternative approach by \citet{van2011lqg} was to use the LQG controller to determine a path distribution, from which the RRT algorithm planned the path by optimizing an objective function based on this distribution. 

Unlike the abovementioned approaches, our method explicitly accounts for the control (motion) uncertainty during the planning phase, paying particular attention to the boundary conditions. 
In contrast to the stabilizing controllers, e.g., Linear Quadratic Regulators and their variants, which rely on linearizing the dynamics and are employed for stabilizing a locally planned trajectory obtained from a sampling-based motion planner~\citep{tedrake2010lqr, van2011lqg, sun2016stochastic},
our method remains dynamics-agnostic and accommodates general robotic dynamics. By exclusively using boundary conditions, we can conveniently construct the framework and incorporate both collision avoidance and goal-reaching constraints. Notably, our approach exhibits remarkable flexibility, allowing us to identify safety-critical subspaces within the state space through a considerably simplified process.

Model Predictive Control (MPC), also known as receding horizon control, represents an alternative strategy for addressing motion uncertainty~\citep{rawlings2017model, bertsekas2005dynamic}. 
In each time step, it tackles a finite horizon optimal control problem and implements the initial action from the computed open-loop trajectory online. 
To reduce computational complexity, MPC commonly utilizes a deterministic model for predicting future states~\citep{bertsekas2012dynamic}. 
Robust MPC~\citep{bemporad1999robust} explicitly addresses model uncertainty by optimizing feedback policies instead of open-loop trajectories. Nevertheless, this optimization process comes with a significant computational burden. Consequently, in practical applications, Tube MPC is often employed as an approximate solution to Robust MPC, focusing only on a specific subspace~\citep{langson2004robust}. 
Tube MPC for nonlinear robot motion models continues to be an active area of research. For instance, the use of sum-of-squares programming~\citep{majumdar2013robust, majumdar2017funnel} and reachability analysis~\citep{althoff2008reachability, bansal2017hamilton} has been explored to solve the nonlinear Tube MPC problem. 
A more recent development is the Model Predictive Path Integral (MPPI) control~\citep{kappen2012optimal, theodorou2010generalized, williams2017information, theodorou2010generalized, williams2018robust}, which is a sampling-based MPC approach for solving finite-horizon stochastic optimal control problems with arbitrary cost functions and nonlinear system dynamics.

\subsection{Markov Decision Process and Value Function Approximation}
Motion planning and control under uncertainty can be generally modeled as a Markov Decision Process (MDP)~\citep{bertsekas2012dynamic, sutton2018reinforcement}, which also serves as a fundamental basis in the proposed framework. 
However, finding the optimal solution for the MDP in real-world applications can pose challenges, primarily stemming from the necessity to store and evaluate an infinite number of points
~\citep{kober2013reinforcement, sutton2018reinforcement, bellman2015adaptive}.  

Since the robot moves in the continuous space, it is therefore also necessary to consider the formulation of general continuous-state MDPs and their solutions. 
The key to solving such MDPs is to obtain the optimal value function which, 
ideally,  should accurately distinguish the boundaries between safe and unsafe regions, subject to robot dynamics constraints.
However, oftentimes real-world problems require approximation to handle the continuous state space in motion planning. This poses a significant challenge on choosing a suitable approximation method that possesses accurate boundaries of the optimal value function and guarantees desired robotic motions around boundaries.

Discretization-based methods, which divide the continuous state into discrete regions and assign approximate transition probabilities to only the neighboring states, can only provide a ``high-level" motion guidance policy~\citep{pereira2013risk, otte2016any, fu2015sense, al2013wind, baek2013optimal}. A more advanced discretization technique is adaptive discretization~\citep{gorodetsky2015efficient, Liu-RSS-19, munos2002variable, lavalle2006planning}. 

An alternative approach may build upon value function approximations, such as neural networks~\citep{fan2020theoretical, hessel2018rainbow, haarnoja2018soft, schulman2017proximal}}, 
kernel methods~\citep{deisenroth2009gaussian, xu2007kernel}, or the combination of linear functions~\citep{munos2008finite, yang2019sample}, to avoid explicit discretization without using a simplified transition function. 
For example, extensive research has focused on approximating the value function using linear combinations of basis functions or parametric functions~\citep{bertsekas1996neuro, sutton2018reinforcement, munos2008finite}. 
The weights in these combinations can be optimized by minimizing the Bellman residual~\citep{antos2008learning}. 
However, these methods face challenges when applied to complex problems due to the non-trivial task of selecting an appropriate set of basis functions. 

To address the above limitation, kernel methods have emerged as a potential solution~\citep{hofmann2008kernel}. By representing the weights in a linear combination of basis functions as products through the dual form of least squares~\citep{shawe2004kernel}, these products can be effectively replaced by kernel functions defined on reproducing kernel Hilbert spaces. 
Once the value functions at supporting states are obtained, the approximation for the value function at any state can be determined. For instance, \citet{deisenroth2009gaussian} employed a Gaussian process with a Radial Basis Kernel (RBF) to approximate the value function. Additionally, there is a significant body of literature on kernelized value function approximations in reinforcement learning~\citep{engel2003bayes, kuss2004gaussian, taylor2009kernelized, xu2007kernel}.
However, these approximations usually cannot generate desirable sharp changes in the value function that correspond to the boundaries between safe and unsafe regions, potentially resulting in collisions.

There is a trend of research on data-driven approaches to obtain the value function approximation. For example, deep learning is used to efficiently computing reachability sets in high-dimensional spaces, which is vital for safe control and navigation in complex environments~\citep{bansal2021, dawson2023}.
Adversarial actor-critic approach is designed to ensure safety in control and reinforcement learning tasks~\citep{ fisac2019, hsu2023}. 
Statistical approximation approaches such as locally weighted projection regression and 
polynomial mixture of normalized Gaussians are also introduced 
in the context of model predictive control~\citep{zhong2023value}.

Although the empirical success of the application of these methods in the context of deep reinforcement learning has been shown in simulated environments or environments without safety considerations~\citep{pan2018reinforcement, pereira2020challenges}, 
applying these methods to safety-critical platforms such as mobile robots is still rare.
One important reason is that the value function representation (either computed through a model or learned by sampling) of these methods cannot strictly satisfy the boundary conditions in the safety-critical regions like obstacles, as these boundaries require special attention -- the value function should not increase towards non-navigable boundaries to ensure safety.
\citet{Xu-RSS-20} show that without careful reward engineering, the value function approximated using neural networks cannot guarantee safety.
This behavior has also been observed in similar works that use neural networks to impose boundary conditions to solve partial differential equations~\citep{raissi2019physics,sukumar2022exact}.
Our work addresses exactly this issue by constructing a safety-aware value function that allows the robot to operate in environments where navigable space is irregular and cluttered.

\subsection{Safety Guarantee with Hamilton-Jacobi Reachability Analysis}
Another closely related field is {Hamilton-Jacobi (HJ) reachability analysis,} which has been applied extensively to certify and ensure the safety of robotic systems~\citep{bansal2017hamilton}.
{It formulates the safety-critical planning problem in terms of a two-player differential game, whose core is to solve the robust version of a first-order HJ partial differential equation (known as the Hamilton-Jacobi Isaacs (HJI) equation)} via the Level Set Method~\citep{mitchell2008flexible}. 
However, it suffers from the curse of dimensionality due to the discretization procedure required by the Level Set Method. 
Many existing works alleviate this problem by decomposing the system into multiple subsystems~\citep{chen2016fast, chen2017exact} or 
by approximating the nonlinear system by a linear time-varying one~\citep{seo2022HJR}. 
Besides solving the HJB equation to find the reachable set via the Level Set Method, another commonly used approach is Sum-of-Squares (SOS) programming.
For example, in~\cite{majumdar2017funnel} and~\cite{kousik2020bridging}, the authors use SOS to compute a set of reachable sets offline and utilize them online to ensure real-time execution.

Two key differences exist between our method and the existing methods in HJ reachability.
First, HJ reachability analysis generally optimizes over the worst-case scenario, i.e., a min-max formulation, to handle the stochastic disturbance, which may lead to conservative behavior~\citep{chen2018hamilton}.
Instead of considering the worst-case disturbances, our work accounts for the system's stochasticity via the 
\textit{probabilistic transitions}~\citep{bellman2015adaptive}. 
To achieve this, we leverage the diffusion approximation to the Bellman equation~\citep{Xu-RSS-20}, which converts the original equation to a second-order Hamilton-Jacobi-Bellman (HJB) PDE and only requires the first two statistical moments of the probabilistic motion model.
The approximation directly considers the uncertainty of the robot's dynamics via the second moment, resulting in less conservative behavior. Its optimal solution is characterized by a second-order nonlinear PDE, a higher order HJB equation in comparison to the first-order HJI equation used in HJ reachability analysis. 

The second significant difference lies in how we tackle the nonlinear HJ-type PDE, which constitutes the core of the problem.  
Most existing studies on HJ reachability rely on a direct utilization of existing solvers, for example, those based on finite difference in combination with the Level Set Scheme~\citep{mitchell2008flexible}. 

In contrast, we present a fresh perspective by proposing a novel PDE computational solution that leverages the Finite Element Method (FEM) and synergizes this with a kernal method. By employing this framework, we aim to enhance the accuracy and efficiency of the solution process.
 Our approach seeks to break new ground and offer an alternative, more advanced solution method. 
It allows for the decomposition of the system into critical and non-critical subspaces, thereby enhancing computational efficiency. 
While prior studies have leveraged a similar system decomposition concept to reduce computation times~\citep{chen2016fast, chen2017exact}, they generally also leverage the commonly-used finite difference Level Set scheme, requiring a case-by-case analysis of the robot dynamics.

\section{Preliminaries}

\subsection{Markov Decision Processes}

Our stochastic planning method is extended from an infinite horizon discrete-time discounted Markov Decision Process (MDP). 
The basic form of MDP is defined by a 5-tuple 
$\langle\mathcal{X}, \mathcal{A}, T, R, \gamma\rangle$,
where $\mathcal{X} \subseteq \mathbb{Z}^{d}$ with $x\in\mathcal{X}$ represents the $d$-dimensional discrete state space, and
$\mathcal{A}$ with $a\in\mathcal{A}$ denotes a finite set of actions; $\gamma \in (0, 1)$ is a discount factor. As an example, a state $x$ may include robot's 2-dimensional position and orientation. 
A robot transits from a state $x$ to the next $x'$ by taking an action $a$ in a stochastic environment and obtains a reward $R(x, a)$. 
Such dynamic transition is governed by a transition function, a conditional probability distribution $T(x, a, x') \triangleq p(x'|x, a)$. The reward $R(x, a)$, a scalar value, 
specifies the one-step reward return that the robot receives by taking action $a$ at state $x$.
We consider the class of deterministic policies $\Pi$, which defines a mapping $\pi \in \Pi: \mathcal{X}\rightarrow \mathcal{A}$. 
Our goal is to maximize the expected discounted cumulative reward $v^{\pi}(x)$, the {\em state value function} of the policy $\pi$ starting at $x$,
\begin{align}\label{eq:value-fn-policy}
    v^{\pi}(x) = \mathbb{E}\left[\sum_{t=0}^{ \infty} \gamma^{t} R(x_t, \pi(x_t))|x_0 = x\right]. 
\end{align}
The state at the next time step, $x_{t+1}$,  draws from distribution $p(x_{t+1} | x_t, \pi(x_t))$. 
The above equation can be written recursively and expressed as the {Bellman equation}
\begin{equation}\label{eq:value-function}
\begin{split}
    v^{\pi}(x) =& R(x, \pi(x)) + \gamma\,\mathbb{E}^{\pi}[v^{\pi}(x')|x], 
\end{split}
\end{equation}
where $\mathbb{E}^{\pi}[v^{\pi}(x')|x]=\int p(x'|x, \pi(x))v^{\pi}(x')\,dx'$.
Solving an MDP amounts to finding the optimal policy $\pi^*$ with the optimal value function that satisfies 
\begin{equation}\label{bellman-optimality-equation}
  v^{\pi^*}(x) = \max_{\pi}\left\{R(x, \pi(x)) + \gamma\,\mathbb{E}^{\pi}[v^{\pi}(x')|x]\right\}.
\end{equation}

\subsection{Necessity of Continuous State MDP Formulation in Motion Planning Tasks}\label{sec:necessity-of-cont}
Using the discrete state MDP to describe motion control problems requires tessellating the robot workspace into grids with each grid being viewed as one state. 
This grid-based discrete modeling has two big problems.
First, it will force the transition probability functions to 
{be discretized from the continuous counterpart} and defined in an ad-hoc jumping fashion. 
However, the robot's physical constraints, e.g., non-holonomic dynamics, may cause a significant error in the assumed transition. 
When sharp-cornered obstacles or sharp-edged holes are close to the robot, the MDP control policy may lead to collision or stuck situations.
Similar approaches such as abstraction-refinement~\citep{junges2022abstraction} construct abstract, coarse MDPs, e.g., by discretization~\citep{huynh2016incremental}, which are easy to solve.
The solution to the abstract MDPs is used to bootstrap the solution to the original MDPs to increase the computational speed.
Regardless of the abstraction method used, our modeling and algorithm framework can serve as the base subroutine to construct the solution to all these approaches.

Second, even if we model the real-world robot dynamics in the discrete representation,
it also has an inherent drawback due to the discontinuity in the value function. 
Depending on the discretization resolution and the magnitude of actions, the robot's next state may not be able to ``escape" from the current grid.  
Specifically, since the value function is the same in the same grid, this 
can prevent state-values at the goal or obstacles from propagating over the entire state space, resulting in a flat value function in most regions. 
Thus, 
the value function 
might fail to capture and represent the underlying objectives (i.e., guidance towards goal and away from obstacles). 
Though one may use reward-shaping techniques (e.g., adding a distance function to each state) to overcome this difficulty, 
the engineering bias towards the solution can be inevitably embedded in the final policy,  often leading to sub-optimality~\citep{riedmiller2018learning}. 

In contrast, a continuous state model does not have the aforementioned issues because we can naturally model the change in the value function at any point in the state space.
Thus, we opt to leverage the continuous state model in our work. 

\subsection{Diffusion MDP (DMDP) Approximates MDP in the Limit} 
We choose to employ a more appropriate continuous state space model that extends beyond the conventional MDP. We offer a lucid and succinct rationale for this extension as follows.

We begin with an MDP of discrete states. By increasing the number of states and letting the distance between any neighbored discrete states become smaller and smaller, one can deduce that it will eventually become an MDP of continuous states. This is equivalent to thinking of a series of MDPs with discrete states closer and closer to each other. 
Then this series of MDPs will converge to a continuous state MDP in the limit. Accordingly their discrete transition distributions will converge to the real continuous transition distribution as well.  
Formally, we assume continuous state $x\in\mathbb{R}^d$ and 
consider an infinitesimal time step at state $x$ in Eq.~\eqref{eq:value-function}. Within such an almost zero time interval, the robot's state can merely shift a little from the current state. It is then reasonable to assume $v^{\pi}(x')$ only changes from $ v^{\pi}(x)$ by a small amount if the value function is smooth. 
Suppose that the value function $v^{\pi}(x)$ for any given policy $\pi$ has continuous first and second order derivatives. We can expand $v^{\pi}(x')$ around $x$ up to the second order as
\begin{align*}
    v^{\pi}(x) + (\nabla v^{\pi}(x))^T(x' - x) +\frac{1}{2}(x' - x)^T\nabla^2v^{\pi}(x)(x' - x),
\end{align*}
where the operator $\nabla$ is the gradient operator and $\nabla^2v$ denotes the Hessian matrix.
Plug this expression of $v^{\pi}(x')$ into  Eq.~\eqref{eq:value-function}, then the Bellman equation reads
\begin{align}
    &-R(x, \pi(x)) \nonumber\\
   &= \gamma\,\Big((\mu^{\pi}_x)^T\,\nabla v^{\pi}(x) +\frac{1}{2}\nabla\cdot \sigma_x^{\pi}\nabla v^{\pi}(x)\Big) - (1-\gamma)\,v^{\pi}(x), \label{bellman-type-pde}
\end{align}
where the notation $\cdot$ in the last equation indicates an inner product; $\mu^{\pi}_x$ and $\sigma^{\pi}_x$ are the first moment (i.e., a $d$-dimensional vector) and the second moment (i.e., a $d$-by-$d$ matrix) of transition functions, respectively. 
The operator $\nabla\cdot \sigma^{\pi}_x\nabla$ reads $\nabla\cdot \sigma^{\pi}_x\nabla = \sum_{i,j}(\sigma^{\pi}_{x})_{i,j}\frac{\partial^2}{\partial x_i\partial x_j},$ 
where $(\sigma^{\pi}_{x})_{i,j}$ is the element at the $i$-th row and $j$-the column of $\sigma^{\pi}_{x}$, and $x_i$ denotes the $i$-th element in the vector $x$.
In practical applications, the first and second moments can be estimated using existing available testing/trial data~\citep{xu2023causal}. 

The preceding process essentially employs the {\em diffusion approximation} to the discrete state MDP in the limit in the continuous state space.
Eq.~\eqref{bellman-type-pde} is a diffusion-type PDE and characterizes the above DMDP model. Its role is analogous to the role of Bellman equation in the discrete state MDP. From the above derivation, the DMDP can be considered as a continuous approximation to the MDP in the limit~\citep{braverman2018taylor}. It only requires the first and second moments of actual transitions as opposed to the full but approximate form of the transition model in the discrete states. In this paper, we use the DMDP as a substrate of the continuous state control model,  and develop numerical solutions to the PDE with required boundary conditions for stochastic motion planning and control problems. We next discuss these boundary conditions and their relation to the mission and environmental constraints. 

\subsection{Boundary Conditions of DMDP}
In this work, we formulate the planning problem as a sparse-reward MDP, where the robot's behavior is only shaped by the boundary conditions.
This formulation is intuitive and requires only minimal reward function design. Let $\Omega$ denote 
the robot's workspace, which is a 
closed and bounded
domain in $\mathbb{R}^d$. 
Let $\partial\Omega$ be its boundary that consists of two parts, i.e., $\partial\Omega = \partial\Omega_1 \cup \partial\Omega_2$. 
The first boundary, $\partial\Omega_1$, represents the safe state-space boundary from which the robot is not allowed to depart (e.g., determined by the obstacles or risky areas in the field). 
We first observe that the {\em value function should not increase towards this physical boundary; otherwise, it will result in actions that guide the robot to the risky areas.} 
To achieve this, a relatively simple boundary condition to impose can be that, at any boundary state, its directional derivative of the value function with respect to its outward unit normal $\hat{\bm{n}}$, 
is zero. 
The second boundary condition, $\partial\Omega_2$, is related to tasks, e.g., the goal state setting. 
We set a constant positive reward $\hat{u}$ on the goal boundary, $\partial\Omega_2$.
Additionally, we impose the transition function on the workspace boundary to be absorbing $p(s'|s,a) = \delta(s'-s), \text{ for all } s \in \partial \Omega$.
This property ensures that once the robot reaches the boundary, it cannot transition to other states, thereby preventing it from acquiring positive rewards from the goal boundary.
Thereby, the optimal policy derived from this setting inherently avoids the obstacle boundary and is goal-reaching.
 
Putting these considerations together, one can easily write the boundary conditions of the DMDP in Eq.~(\ref{bellman-opt-boundary-conditions}). Additionally, 
from the derivation in the previous section, the optimal policy and optimal value function should satisfy the following PDE in Eq.~(\ref{nonlinear-diffusion-pde}), analogous to the Bellman optimality equation Eq.~\eqref{bellman-optimality-equation} in MDP :
\begin{eqnarray}\label{eq:strong-form-optimality}
 0&=&\max_{\pi}\Big\{\gamma\,\Big((\mu_x^\pi)^T\nabla v^{\pi}(x) +\frac{1}{2}\nabla\cdot \sigma_x^\pi\nabla v^{\pi}(x)\Big) + \nonumber\\
 &{}& R(x, \pi(x)) - (1-\gamma)\,v^{\pi}(x)\Big\}, 
 \label{nonlinear-diffusion-pde}
\end{eqnarray}
with boundary conditions
\begin{subequations}\label{bellman-opt-boundary-conditions}
\begin{eqnarray}
    \sigma_x^{\pi}\,\nabla v^{\pi}(x)\cdot \hat{\bm{n}} &=& 0, \mbox{  on } \partial\Omega_1, \label{bellman-opt-boundary-condition-a}\\
    v^{\pi}(x) &=& \hat{u},\mbox{  on } \partial\Omega_2,  \label{bellman-opt-boundary-condition-b}
\end{eqnarray}
\end{subequations}
where $\hat{\bm{n}}$ denotes the unit vector normal to $\partial\Omega_1$ pointing outward, and $\hat{u}$ is a constant value. 
Boundary condition in the form of~\eqref{bellman-opt-boundary-condition-a} has been explored and successfully tested when representing the known space using a Laplace equation~\citep{shade2011choosing}.

Note that Eq.~(\ref{nonlinear-diffusion-pde}) is a nonlinear PDE (due to the maximum operator over all policies), and it can be thought of as a type of Hamilton-Jacobi-Bellman (HJB) equation. 
Despite this nonlinearity, our algorithm (a policy iteration procedure) in the future section will allow to solve a linear PDE for a fixed policy and iteratively search for the optimal solution.

\section{Methods}
We propose a principled policy iteration procedure built upon the framework of finite element methods (FEM) to solve the DMDP. The FEM is able to generate solutions that satisfy complex geometric boundary conditions which are safety-critical in motion control applications (Section~\ref{FEM-framework}). 
To overcome the computational challenge brought by the pure FEM, 
we design a basis function space consisting of tensor product of 
kernels and mesh-based polynomials (Section~\ref{tensor-product-basis}). The mesh-based polynomials 
will handle the required boundary constraints in location-relevant state dimensions whereas the kernels can  approximate the integrals of other dimensions. 
We finally integrate our methods in the proposed policy iteration procedure for practical implementation (Section~\ref{policy-iteration-algorithm}).

\subsection{Finite Element Methods for Policy Evaluation}\label{FEM-framework} 
We propose to use FEM to solve the DMDP. 
Specifically, for a fixed policy, we need to obtain the corresponding value function from Eq.~(\ref{nonlinear-diffusion-pde}) and constraints Eq.~(\ref{boundary-condition}). In this case, Eq.~(\ref{nonlinear-diffusion-pde}) becomes the linear PDE shown in Eq.~(\ref{bellman-type-pde}). 
Our focus now turns to {\em efficiently solve this boundary-valued problem}. Our approach is to build on the {\em Galerkin} approach in FEMs~\citep{brenner2008mathematical}.

First, we introduce the following notations to make a compact representation of Eq.~(\ref{bellman-type-pde}) and eliminate the constant $\hat{u}$ on the boundary
\begin{align*}
D(x)\triangleq -\frac{1}{2}\,\gamma\,\sigma_x^\pi;\,\,\,\,\,\,\,\,  &q(x)\triangleq \gamma\,\mu_x^\pi; \\
b\triangleq-(1-\gamma);\,\,\,\,\,\,\,\,  &f(x)\triangleq -R(x, \pi(x)) - (1-\gamma)\hat{u}, 
\end{align*}
where $x\in\mathbb{R}^d$. 

By definition, $D(x)$ is a $d$-by-$d$ symmetric positive semi-definite matrix, $q(x)$ is a vector of dimension $d$, $b$ is a constant, and $f(x)$ is a scalar function.
As $\hat{u}$ is a constant, 
we can substitute $v(x)-\hat{u}$ with a new $v(x)$ into 
Eq.~(\ref{bellman-type-pde})
with constraints Eq.~(\ref{boundary-condition}), and this yields
\begin{align}\label{method:diffusion-pde}
-\nabla\cdot D(x)\nabla v(x) + q^T(x)\,\nabla v(x) + b(x)\,v(x) = f(x), 
\end{align}
with boundary conditions
\begin{subequations}\label{boundary-condition}
\begin{eqnarray}
   D(x)\,\nabla v(x)\cdot \hat{\bm{n}} &=& 0, \mbox{  on } \partial\Omega_1, \label{boundary-condition-a}\\
    v(x) &=& 0, \mbox{  on } \partial\Omega_2. \label{boundary-condition-b}
\end{eqnarray}
\end{subequations}
{The operation of $v(x)-\hat{u}$ merely translates the original function by $\hat{u}$, and accordingly the solution is simply shifted by $\hat{u}$ as well from the original solution.}
In the framework of the Galerkin approach, 
we aim to find the solutions to satisfy its variational form, also known as the {\em weak form}~\citep{hughes2012finite}. The variational/weak form of the problem amounts to multiplying both sides of Eq. (\ref{method:diffusion-pde}) by a sufficiently smooth {\em test function} $\omega(x)$ with constraint $w(x)=0$ on $\partial \Omega_2$. 
By the integration-by-parts formula and further simplifying the function notation, we have
\begin{align*}
    &-\int_{\Omega}\omega\nabla\cdot D\nabla v\,dx \\
    &= \int_{\Omega}\nabla \omega\cdot D\nabla v \,dx + \int_{\partial\Omega_1} \omega\,D\,\nabla v\cdot \hat{\bm{n}}\,ds.
\end{align*} 
Because of the boundary condition Eq. (\ref{boundary-condition-a}), we can transform Eq.~\eqref{method:diffusion-pde} into the variational formulation
\begin{align}
    \int_{\Omega} \left(\nabla \omega\cdot D\nabla v + w\,q^T\,\nabla v + b\,\omega\, v- f\,\omega\right)dx = 0.
    \label{weak-diffusion-pde}
\end{align} 
One can immediately see the \textit{advantages} of the variational form: it explicitly uses the boundary conditions Eq.~(\ref{boundary-condition-a}), and it only requires the first derivative of value function rather than the second derivative in the original PDE. 
Moreover, any solution to Eq. (\ref{method:diffusion-pde}) automatically satisfies this variational form Eq. (\ref{weak-diffusion-pde}). This is because as described above, Eq. (\ref{weak-diffusion-pde}) is obtained through multiplying both sides of Eq. (\ref{method:diffusion-pde}) by a test function. It can be shown the solution to this variational form is also the solution to the original form under sufficient smooth conditions~\citep{oden2012introduction}. 

Now we ``mesh" the domain $\Omega$ into suitable discrete elements, and denote the domain consisting of all elements by $\Omega^h$.
An example of mesh in a $2$-dimensional space consists of 
triangle elements which can effectively approximate any geometric regions. 
Our aim is to find an approximate solution to the form
\begin{align}
    v^h = \sum_{i=1}^N a_i \phi_i(x), \,\,\,\,\,\,\mbox{with $v^h=0$ on $\partial\Omega_2$,} 
    \label{approx-v}
\end{align}
where $\phi_i(x)$ are called basis functions, each of which is defined only on a limited number of elements, and $\phi_i(x)=0$ on other elements. 
{
The basis functions of FEM are crucial in ensuring that the value function is accurately approximated and that the boundary conditions are satisfied by construction. 
Two key properties of these basis functions contribute to these desirable outcomes. 
First, their compact support, meaning that each basis function only interacts with a limited number of neighboring basis functions, leads to a local representation of the value function. 
This allows for a highly accurate description of local variations. 
Secondly, the basis functions are continuous across individual elements, 
and the continuity 
is instrumental in accurately representing continuous sharp gradients along boundaries, especially in scenarios where precision near boundaries is vital, such as safety-critical problems.

In practice, basis functions are often constructed using interpolation polynomials defined at nodes of adjacent elements (as detailed in Section 4 of \citet{oden2012introduction}).

The test function 
can also be
a combination of the same basis functions, i.e., $\omega^h(x)=\sum_{i=1}^N c_i\phi_i(x)$ and 
require 
$\omega^h(x)=0$ on $\partial\Omega_2$. 
After substituting $\omega^h(x)$ into Eq.~\eqref{weak-diffusion-pde}, 
we can deduce a system of linear algebraic equations, 
\begin{align}
    KA = F,
    \label{stiffen-eqn}
\end{align}
where $A=(a_1, \ldots, a_N)^T$ with each $a_i$ being an coefficient that needs to be estimated.
Each entry of matrix $K$ ($\triangleq [K_{ij}]_{N\times N}$) corresponds to basis function involved integrals on the right-hand side of Eq. (\ref{weak-diffusion-pde}):
\begin{align}
    K_{i,j}=\int_{\Omega^h} \left(\nabla\phi_i\cdot D\nabla\phi_j + \phi_i\,q^T\,\nabla\phi_j + b\,\phi_i\,\phi_j\right)\,dx, 
    \label{stiffen-K}
\end{align}
and elements in $F$ ($\triangleq (F_1,\ldots, F_N)^T$) correspond to the products with $f$: 
\begin{align}
    F_{i}=\int_{\Omega^h} f\,\phi_i\,dx.
    \label{stiffen-F}
\end{align}
Because the support of every basis function only includes a limited number instead of all elements, every integral is carried out on a limited number of elements as well. All integrals above are evaluated numerically based on interpolation polynomial basis. 
Solving the linear system Eq.~(\ref{stiffen-eqn}) gives the estimates of $a_i$. Accordingly, we obtain the approximate value function $v^h(x)$.

\subsection{Tensor product of kernel and polynomial basis functions}\label{tensor-product-basis}

Despite its broad applications, one of the major computational hurdles of the mesh-based FEM is numerical integration over each element. 

On the one hand, it is well recognized that meshless methods such as kernel-based quadrature rules are 
computationally faster for numerical integrals 
~\citep{wendland2004scattered, bach2017equivalence}. 
On the other hand, we observe that only a small number of state dimensions are critical to boundary constraints in most robotic motion planning and control problems. For example, violating the obstacle boundary conditions in {\em location related} state components is strictly prohibited when robots operate in cluttered obstacle environments. 
Safety in other state components such as the robot's orientation can be handled by alternative methods, e.g., obstacle inflation~\citep{marder2010office}.

Therefore, we propose a tensor product of kernel and interpolation polynomial basis functions in the principles of Galerkin methods. 
The proposed method can 
satisfy the boundary conditions in boundary-critical state dimensions (e.g., location-relevant) up to the mesh element resolution --- an advantage of finite elements, and meanwhile reduce the computation load of numerical integrals --- an advantage of the kernel approximation to integrals.

Formally, we divide state variable $x = (x^q, x^p)^T\in\mathbb{R}^d$ where $x^q=(x_1,\ldots, x_{d_1})^T\in\mathbb{R}^{d_1}$ is the non-boundary-critical state components and $x^p=(x_{d_1+1},\ldots,x_{d_1+d_2})^T\in\mathbb{R}^{d_2}$ is boundary-critical state components (e.g., location related) components. Obviously, $d=d_1+d_2$. (Note, the notations $x^q$ and $x^p$ should not be confused with a variable of exponent.) 
We construct a set of tensor products of kernel and interpolation polynomial
\begin{align}\label{set:tensor-product}
\left\{\,v^h_{i,t}(x)\triangleq k(x^q, \xi_t)\phi_i(x^p), i=1,\ldots, N_1, t=1,\ldots, N_2\right\}.
\end{align}
By some abuse of notations, $\phi(\cdot)$ denotes the (linear Lagrange) interpolation polynomial on $d_2$-dimensional finite elements. $k(\cdot, \xi_t)$ is a positive semidefinite kernel function defined in $\mathbb{R}^{d_1}$, where $\xi_t\in\mathbb{R}^{d_1}$ is a fixed (sampled) point in non-boundary-critical dimensions. Thereafter we call $\xi_t$ the {\em supporting states}. One typical kernel is the Gaussian kernel with the form
\begin{align}\label{Gaussian-kernel}
k(x^q, \xi_t) = c\,\exp\left(-\frac{1}{2\kappa^{2}}(x^q-\xi_t)^T(x^q-\xi_t)\right), 
\end{align}
where $\kappa$ is a scale factor and $c$ is a normalizing constant (although $\kappa$ can be different for differing supporting states, it is not essential in our method development).
{It is also important to note that while we define the kernel lengthscale $\kappa$ as a scalar in Eq. \eqref{Gaussian-kernel} for the notation clarity, our method can be used with more general kernel functions as long as it is second-order differentiable.
In subsequent experiments, we adopt a diagonal matrix representation for $\kappa$ to accommodate variations in units across different state dimensions effectively.}

We view the tensor products as new basis functions and represent the value function in the DMDP by the following form
\begin{align}\label{kernel-poly-v}
    v^h(x) = \sum_{i, t=1}^M a_{i,t}\,k(x^q, \xi_t)\phi_i(x^p) \triangleq \sum_{i, t} a_{i,t}\, v^h_{i,t}(x),
\end{align}
where $M=N_1N_2$. 
Following the exact steps from the variational form of the boundary-value problem in Galerkin methods in Section~\ref{FEM-framework}, we can arrive at a similar linear system of equations $KA=F$. 
In this case, matrix $K$ is $M\times M$; vectors of $A$ and $F$ are of $M\times 1$.

The advantages of the proposed tensor product basis functions lie in the way that the kernel function can {\em simplify the numerical computation of integrals}. 
To see this, here we 
detail the calculations involved in the linear system of equations.
Let us order the tensor product basis by indices $\hat{i}=(i-1)N_2 + t$ in the set (\ref{set:tensor-product}). The element in matrix $K$ then is
\begin{align}\label{tensor-stiffen-K}
    K_{\hat{i}, \hat{j}}=\int_{\Omega^h} \left(\nabla v^h_{i,t}\cdot D\nabla v^h_{j, l} + v^h_{i,t}\,q^T\,\nabla v^h_{j, l} + b\,v^h_{i, t}\,v^h_{j, l}\right)\,dx,
\end{align}
where similarly to $\hat{i}$, $\hat{j}$ is understood as $jN_2+l$ with $l=1,\ldots,N_1$. The derivative of each tensor product basis with respect to $x$ can be written as
\begin{subequations}
\begin{align*} 
   \nabla v^h_{i,t} 
     &=  \left( \phi_i(x^p)\nabla_{x^q} k(x^q, \xi_t),  k(x^q, \xi_t)\nabla_{x^p}\phi_i(x^p) \right)^T\\
     &\triangleq \left(\nabla_{x^q}v^h_{i, t}, \nabla_{x^p}v^h_{i, t}  \right)^T. 
\end{align*}
\end{subequations}
The above notations are used to separate the $x^q$ and $x^p$ components.  
Let $D = \begin{bmatrix}
D_{qq}, & D_{qp}  \\
D_{pq}, & D_{pp}
\end{bmatrix}$,
where $D_{qq}$ is of $d_1\times d_1$, $D_{qp}$ ($=D_{pq}^T$) of $d_1\times d_2$, and $D_{pp}$ of $d_2\times d_2$,
then we have the following expression for the first term in Eq.~(\ref{tensor-stiffen-K})
\begin{align}\label{diffusion-term-tensor-basis}
    &\nabla v^h_{i,t}\cdot D\nabla v^h_{j,l} =  \nonumber\\
    &\nabla_{x_q}v^h_{i, t}\cdot D_{qq} \nabla_{x^q}v^h_{j, l} + \nabla_{x^q}v^h_{i, t}\cdot D_{qp} \nabla_{x^p}v^h_{j, l} \nonumber\\
    &+ \nabla_{x^p}v^h_{i, t}\cdot D_{pq} \nabla_{x^q}v^h_{j, l} + \nabla_{x^p}v^h_{i, t}\cdot D_{pp} \nabla_{x^p}v^h_{j, l}.
\end{align}
We can doubly integrate these terms, where an inner integral is with respect to $x^q$ and an outer integral is with respect to $x^p$. 
Specifically, one can evaluate the inner integral with the properties of kernels, or get a closed-form approximation using kernels; The outer integral can be numerically computed by the standard methods for integration polynomials. 

We use the integral of the first term in Eq.~(\ref{diffusion-term-tensor-basis}) that involves $D_{qq}$ to illustrate the above point. 
We write out the expression of integral
\begin{align}\label{diffusion-term:first-term-integral}
    &\int_{\Omega^h} \nabla_{x^q}v^h_{i, t}\cdot D_{qq} \nabla_{x^q}v^h_{j, l}\,dx 
    =\int_{\Omega^h_p} \hat{D}_{qq}\phi_i(x^p)\phi_j(x^p)\,dx^p,
\end{align} 
where 
\begin{align}
   \hat{D}_{qq} = \int_{\Omega^h_q}\nabla_{x^q} k(x^q, \xi_t)\cdot D_{qq}\nabla_{x^q} k(x^q, \xi_l)\,dx^q.
\end{align}
Notations $\Omega_q^h$ and $\Omega_p^h$ refer to the integrals taken in $d_1$ and $d_2$ dimensions, respectively. As for $\hat{D}_{qq}$, we use an approximation for Gaussian kernels (Eq.~(\ref{Gaussian-kernel}))
\begin{align}
   \hat{D}_{qq} &= c^2\int_{\Omega^h_q} \kappa^{-2}(x^q - \xi_t)^T\cdot D_{qq} \kappa^{-2}(x^q - \xi_l) \nonumber\\
   &\exp\left(-\frac{1}{2\kappa^{2}}( 2\lVert x^q - \frac{\xi_t + \xi_l}{2} \rVert^2 + \frac{1}{2}\lVert \xi_t -\xi_l\rVert^2)\right)\,dx^q\nonumber\\
   &\sim_{c,\kappa} (\xi_l - \xi_t)^T\cdot D_{qq} (\xi_l - \xi_t) \exp\left(-\frac{1}{4\kappa^{2}}\lVert \xi_t -\xi_l\rVert^2\right),
\end{align}
where $D_{qq}$ should be evaluated at $\frac{\xi_t +\xi_l}{2}$ in the last equation, and notation $\sim_{c,\kappa}$ means that we only keep the main term for simplicity by ignoring all constant multipliers. The smaller $\kappa$, the more accurate the approximation.

From the above derivation, the introduction of kernels as a part of tensor product basis is clear: it serves to simplify or approximate the integration on higher dimensions (i.e., $d_1$ dimensions) in the solution framework.
As noted before, the outer integral with respect to $x^p$ in Eq.~(\ref{diffusion-term:first-term-integral}) is numerically computed on finite elements in $d_2$-dimensional space, i.e., the boundary-critical state dimensions. Since  $d_2$ is typically small, the computation can be carried out efficiently.

To save space, we omit the details of integrals of other terms in Eq.~(\ref{diffusion-term-tensor-basis}), and put them in the Appendix~\ref{app:stiffen-matrix-elements} (titled Appendix to Section~\ref{tensor-product-basis}). 
In addition, we provide the computation of $F$ using Gaussian kernel as the approximation that resembles a reproducing function property $ F_{\hat{i}}\approx \int_{\Omega_p^h} f(\xi_t, x^p)\phi_i(x^p)\,dx^p.$ 
Once we compute the matrix $K$ and vector $F$, the solution of the linear system $KA=F$ gives $a_{i,t}$ in Eq.~(\ref{kernel-poly-v}). Thus the value function with a given policy is finally obtained.  

It is worth mentioning that, we presented our method for the general case in this section. In Appendix~\ref{appendix:3d-construction} (titled Construction of Kernel and Polynomial Tensor Product Basis in 3-dimensional State Space), 
we provide detailed computation steps for a case in $3D$ space.

\subsection{Policy Update}~\label{sec:policy-improvement}
Based on the current value function computed from the step of policy evaluation, we can then perform the policy update step to improve the action on each state.   
Since the linear interpolation polynomial is applied to construct the tensor product basis function, we cannot directly use the DMDP's optimality equation (Eq.~\eqref{eq:strong-form-optimality}) as it requires a second-order derivative of the value function. Instead, 
we use the original Bellman optimality equation for the policy update
\begin{equation}\label{eq:bellman-optimality}
    \pi(x) = \argmax_{a \in \mathcal{A}}\Big\{R(x, a) + \gamma \mathbb{E}\big[v^{\pi}(x')|x'\sim p(x'|x,a)\big] \Big\}.
\end{equation}
Evaluation of the expectation $\mathbb{E}[v^{\pi}(x')]$ requires an unknown transition function $p(x'|x,a)$. To approximate the transition probability function, we leverage the first two moments in DMDP to construct a Gaussian distribution: $x'_{i} \sim \mathcal{N}(x' | \mu_x^{a}+x, \sigma_x^{a}-\mu_x^{a}(\mu_x^{a})^{T})$.
Then, the expectation is estimated as $\mathbb{E}[v^{\pi}(x')] \approx \frac{1}{m}\sum_{i=1}^{m}v^{\pi}(x'_{i})$, where $x'_{i}$ are sampled from the Gaussian distribution. 
Finally, we choose the best action according to Eq.~\eqref{bellman-optimality-equation}.

\subsection{Policy Iteration Algorithm}\label{policy-iteration-algorithm}
\begin{algorithm}[t] 
\caption{{Policy Iteration with Kernel and Finite Elements}}
\label{alg:pi}
\begin{algorithmic}[1]
    \INPUT{
    A set of elements in the boundary-critical state dimensions 
    and node points on elements $\mathbf{X}^{p}=\{x^p_{1}, ..., x^{p}_{N_1}\}$;
    the feasible space and goal boundaries; 
    a set of supporting states for kernel functions in non-boundary-critical state dimensions $\mathbf{\xi} = \{\xi_1, ... \xi_{N_2}\}$;
    the kernel function $k(\cdot, \cdot)$; 
    the DMDP's moments and reward function $\mu, \sigma, R$;
    }
    \OUTPUT{The coefficients $A = \{a_{i,t}|i = 1...N_1, t = 1...N_2 \}$ of the optimal value function.}
    \State Construct the 
    linear interpolation polynomial functions $\phi_i, i=1...N_1$ on elements.
    \State Initialize the action on the states $\mathbf{X} = \{x^p_{i} | i = 1...N_1\} \times \{\frac{\xi_t + \xi_l}{2} | t=1...N_2, l=1...N_2\}$.
    \Repeat
        \State // Policy evaluation step
        \State Construct the matrix $K$ and the load vector $F$ based on Section~\ref{tensor-product-basis} 
        \State Solve the linear system $KA = F$ for the coefficients.
        \State // Policy improvement step
        \For {$x_i \in \mathbf{X}$} 
            \State Update the action at the state $x_i$ based on Eq.~\eqref{eq:bellman-optimality}. 
        \EndFor
    \Until Change in coefficients $A$ is sufficiently small
  \end{algorithmic}
\end{algorithm} 

The whole algorithm is pseudo-coded in Alg.~\ref{alg:pi}.
The algorithm requires the mesh information of the boundary-critical dimensions, including the mesh configuration and the boundary areas.
The mesh configuration can be obtained by any meshing method, e.g., Delaunay triangulation~\citep{cheng2013delaunay}.
The goal boundary is typically defined by the task.
The boundaries can be obtained from the perception result, e.g., segmenting navigable and non-navigable spaces from camera or LiDAR sensors.
The algorithm computes the optimal value function and the policy by iterating over the policy evaluation and improvement steps until the coefficients in constructed solution stop changing. 

In Appendix~\ref{appendix:Method-Analysis} (titled Analysis of the Proposed Approach),  
we analyze the stability of our method. We show that the value function in the proposed approach consists of two desired properties. One is that the maximum of the value function is at the goal and no local maximum inside the feasible motion region for our proposed design of rewards and goal values. The other is that the derived trajectories of the robot always lead to the goal without leaving the safe region
under mild conditions on the robotic dynamics model. The proposed hybrid representation of the value function thereby preserves these two advantages. We refer the readers to Appendix~\ref{appendix:Method-Analysis} for these conditions and detailed derivations.

\begin{table*}[t]
\caption{Success and Collision Rates of Different Methods for Varying Obstacle Ratios. The best success and collision rates for each obstacle ratio are highlighted in \textbf{bold}.}
\centering
\begin{tabular}{c|ccccc|ccccc}
\hline\hline
& \multicolumn{5}{c|}{Success Rate} & \multicolumn{5}{c}{Collision Rate} \\
\hline
 & 5\% & 10\% & 15\% & 20\% & 25\% & 5\% & 10\% & 15\% & 20\% & 25\% \\
\hline
2D Mesh + 1D Kernel (Ours) & \textbf{1} & \textbf{1} & \textbf{0.98} & \textbf{0.91} & \textbf{0.86} & \textbf{0} & \textbf{0} & \textbf{0} & \textbf{0} & \textbf{0} \\
3D Mesh (Ours) & \textbf{1} & \textbf{1} & 0.90 & 0.89 & 0.84 & \textbf{0} & \textbf{0} & \textbf{0} & \textbf{0} & \textbf{0} \\
3D Kernel~\citep{xu2007kernel}~\citep{Xu-RSS-20} 
& \textbf{1} & 0.88 & 0.42 & 0.36 & 0.39 & \textbf{0} & 0.01 & 0.16 & 0.29 & 0.42 \\
Grid~\citep{thrun2000probabilistic} & 0.91 & 0.78 & 0.45 & 0.43 & 0.15 & 0.01 & 0.03 & 0.15 & 0.21 & 0.23 \\
\hline\hline
\end{tabular}
\label{tb:success-collision-rate}
\end{table*}

\section{Simulated Experiments}
We evaluate our method for mobile robot navigation through multiple simulations of increasing realism.
The first simulation uses a 3-dimensional-state motion model to plan and simulate (test) the method's performance.
Using this experiment, we aim to evaluate the method's ability to guarantee safe and goal-reaching navigation under uncertain dynamics. 
The second simulation employs a 4-dimensional model that includes velocity to provide insight into the advantage of the hybrid method.
We then conduct the third simulation with a 7-dimensional model involving even higher order derivatives of state variables. 

\subsection{Evaluation with Simplified UGV Model}

\subsubsection{Setup}

We compare the proposed methods with two standard baseline methods for solving the MDP in a $4m \times 4m$ environment with randomly generated obstacles.
The robot is modeled as a discrete-time Dubin's car where 
 its state space is the pose $(\mathrm{x},\mathrm{y},\theta)$ and action space consists of linear and angular velocities, $a=(v,\omega)$.
 (This also prepares our real robot evaluation next which has the same model.)
 
To simulate the action uncertainty, we adopt a popular method~\citep{thrun2002probabilistic} and add
an additive control noise $\epsilon$ to the control input.
Formally, $\mathrm{x}_{t+1}=\mathrm{x}_t+(v+\epsilon)dt\cos{\theta_t}$, $\mathrm{y}_{t+1}=\mathrm{y}_t+(v+\epsilon)dt\sin{\theta_t}$, and $\theta_{t+1} = \theta_{t} + (\omega+\epsilon)dt$, where $dt=0.05s$.
To sample the obstacles, we first discretize the $x,y$ dimensions into $20\times 20$ grids.
We then randomly assign each grid to be an obstacle except for the starting position and the goal position $(3.8, 3.5)$, which are always free. 

The first baseline is the traditional discrete state MDP which discretizes the robot's state space into 3-dimensional cubes, and each cube represents one state~\citep{thrun2000probabilistic}. 
This is the grid-based representation of the value function mentioned in Section~\ref{sec:necessity-of-cont}. 
It uses the same sampling-based policy improvement method mentioned in Section~\ref{sec:policy-improvement}, where the next states are sampled from the robot's stochastic dynamics. 
Note that the grid-based method differs from the mesh-based method in that it requires modification of the transition function as mentioned in Section~\ref{sec:necessity-of-cont}. 
The second baseline uses the kernel method to represent the continuous value function using the DMDP formulation without explicit discretization.
This representation is common in the literature~\citep{deisenroth2009gaussian, xu2007kernel}.
Our method uses rectangular (or cubic) elements with bi-linear Lagrange polynomials in the mesh-based dimensions and the Gaussian kernel for the meshless dimension. 
We evaluate two variants of our method: 2D Mesh + 1D Kernel and 3D Mesh. 
For 2D Mesh, we ensure critical boundary conditions at the $\mathrm{x}-\mathrm{y}$ dimensions and kernel-based representation in the $\theta$ dimension with a $\frac{\pi}{4}$ lengthscale (the scale parameter $\sigma$). 
To make a fair comparison, the full kernel-based method also uses the Gaussian kernel with $(0.2, 0.2, \frac{\pi}{4})$ lengthscale corresponding to each of the dimensions. 
The lengthscales of the kernel method are chosen based on the hyper-parameter search methods described in~\citet{bergstra2012random}. 
We run all the methods until convergence.
All the methods are implemented in C++, and the experiments are conducted using a PC with an 8-core Intel i7 CPU and 16GB RAM.
We use $5$ obstacle ratios $0.05, 0.1, 0.15, 0.2, 0.25$ 
where the obstacle ratio is defined as the number of randomly generated obstacle grids divided by the total number of all grids. This setup might be imagined as a mobile robot maneuvering in a forest where each obstacle grid represents a tree trunk. This scenario is actually very challenging given the confined space and cluttered trees of random placement~\citep{mohamed2022autonomous}.
We randomly generate $30$ environments for each ratio, and each method is tested for $50$ runs.
This equals testing each policy $7500$ times in $150$ different environments.
The run succeeds only if the robot reaches the goal without collision or exceeding $700$ steps.

\subsubsection{Results}
\begin{figure}[t]
   \centering
    \includegraphics[width=0.9\linewidth]{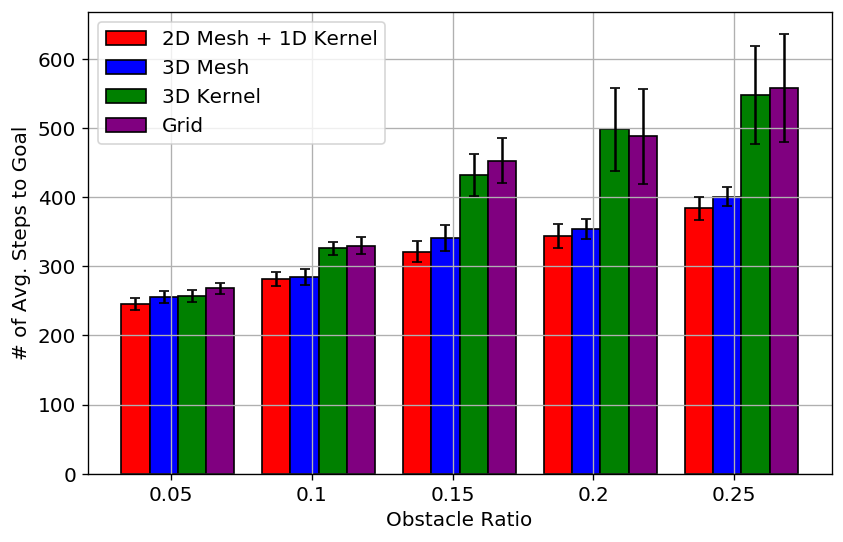}
    \caption{
    The average steps to the goal measure the efficiency of the methods. 
    For the runs that do not reach the goal, including collisions or getting stuck, we set the number of steps as {700}, which is the maximum step of the simulation.
    }
    \label{fig:performance} \vspace{-10pt}
\end{figure}

\begin{figure*}[t]
    \centering
    \subfloat[]{\label{fig:traj-mixed}\includegraphics[width=0.23\linewidth]{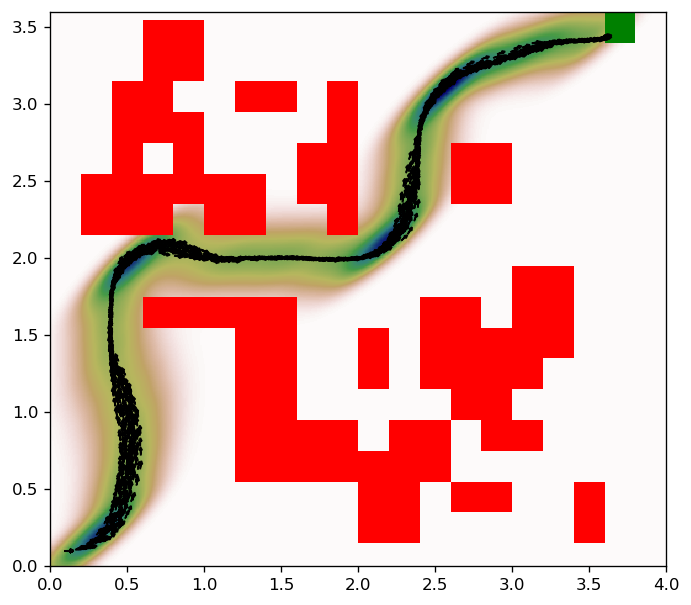}} 
    \subfloat[]{\label{fig:value-mixed}\includegraphics[width=0.23\linewidth]{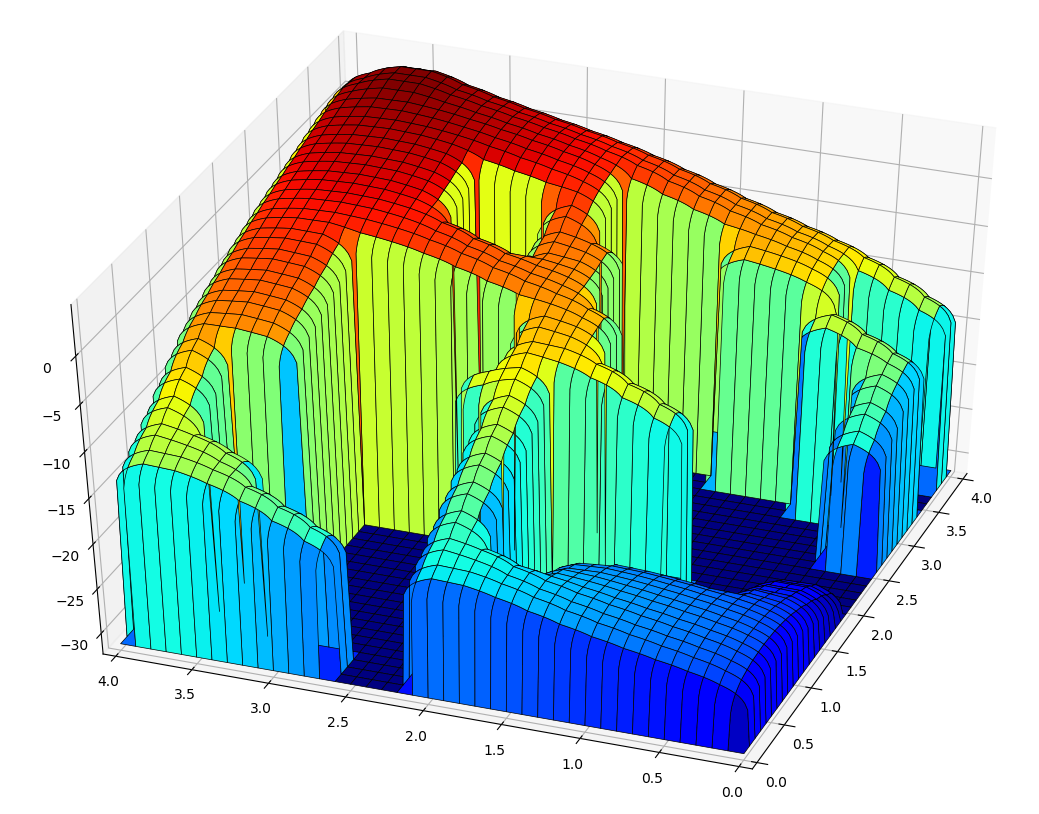}} 
    \subfloat[]{\label{fig:traj-kernel}\includegraphics[width=0.23\linewidth]{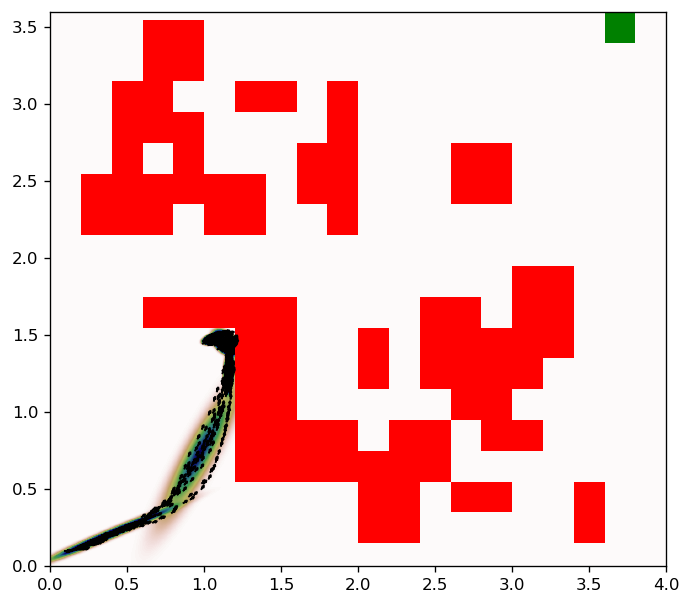}} 
    \subfloat[]{\label{fig:value-kernel}\includegraphics[width=0.23\linewidth]{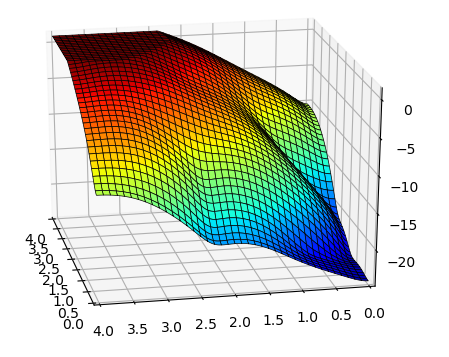}} 
    \caption{
    The multi-trial trajectories of (a) our method (2D Mesh + 1D Kernel) and (c) kernelized value function representation in one $0.25$ obstacle ratio environment demonstrate the ability of our method to navigate in confined spaces despite the motion noise.
    The red and green squares are the obstacles and goals, respectively.
    The black arrows show $50$ trajectories (state-action sequences), and the color gradients around the trajectories visualize the estimated probability density of the robot's trajectory.
    It illustrates the likelihood of the robot visiting each state, given its policy and the stochastic dynamics model.
    Figures (b) and (d) compare value functions in log scale for the $\mathrm{x}-\mathrm{y}$ dimensions at $\theta=\frac{\pi}{2}$ for the two methods. (Please note that the viewing angles of the value function surfaces have been optimized to enhance visualization. In both figures, the upper corners displaying the highest values correspond to the goal regions.)
    }
    \label{fig:trajectories}
\end{figure*}

We first evaluated the success rates and safety properties of the policy computed by each method using the success and collision rates shown in Table~\ref{tb:success-collision-rate}.
They indicate the policy's ability to navigate cluttered environments without getting trapped in a local minimum or colliding with obstacles. 
We can observe that each method's success rate decreases as the environment become more cluttered.
However, the mesh-based methods achieved higher success rates than the grid and kernel methods. 
We also observe that compared with the full mesh-based method (3D Mesh), the hybrid method's (2D Mesh + 1D Kernel) success rate is slightly higher. 
We conjecture that this is the byproduct of the usage of the kernel function in the $\theta$ dimension.
It adds non-linearity to the basis function, whereas FEM uses only linear combination of polynomials to represent the value function.
The additional non-linearity provides our method more flexibility in the value function representation. 
The kernelized (3D Kernel) value function representation method is better than the grid-based method when the environment is not cluttered, i.e., the obstacle ratio is less than $15\%$.
However, its success rate degrades quickly as the number of obstacles increases.
Unsurprisingly, because the mesh-based method can satisfy the safety-critical boundary conditions at the location dimensions, our methods do not have any collision in all the experiments, indicated by the $0$ collision rate in the right part of the table.
In contrast, the kernel method's collision rate rises quickly as the number of obstacles increases. 
Although the proposed methods can achieve zero collision rates, we can observe that their success rate are not $100\%$ in cluttered environments. 
This is because, in some environments, the obstacles are placed in a way that prevents the value from propagating to the robot's starting position.
This issue can be overcome by developing an adaptive discretization strategy, discussed in Appendix~\ref{appendix:value-function} (titled Issues of Value Function Propagation). 
In addition to comparing the ability to satisfy the safe and goal-reaching constraints, we also compare the navigation performance of each method using the number of steps (action taken) to arrive at the goal shown in Fig.~\ref{fig:performance}. 
A smaller number of steps means the policy is more efficient. 
Within the successful trials, the policy computed by our method is the most efficient in reaching the goal, and the policy of the grid-based method takes the longest time.

We also compare our methods and the kernel method in terms of the quality of the value function and the policy trajectories in one of the challenging $0.25$ obstacle ratio environments shown in Fig.~\ref{fig:trajectories}. 
The value function of our method satisfies both the geometric shape of the obstacles shown in the flattened $0$ value areas and the goal condition as indicated by the  highest state-value around the goal region in Fig.~\ref{fig:trajectories}\subref{fig:value-mixed}. 
The resulting policy navigates the robot through the narrow passage and arrives at the goal despite the stochastic control noise, as shown in Fig.~\ref{fig:trajectories}\subref{fig:traj-mixed}. 
In contrast, although the kernelized value function obtains the highest state value at the goal region, it cannot strictly satisfy the state values at the obstacle regions due to the inherited smoothing effect of the kernel function.
Thus, it is natural that the policy obtained from this value function navigates the robot into obstacles in Fig.~\ref{fig:trajectories}\subref{fig:traj-kernel}.
We can also observe that the parameter search reveals a preference for a {\em smooth} value function to {\em generalize} across various obstacle ratios. 
Intuitively, we can interpret such smoothness in terms of ``safety": 
since our target environment has many possibilities of obstacle configurations,  a smaller length scale can capture sharp changes in value functions but meanwhile can also restrict state value propagation during the MDP optimization process, leading to inferior value function approximation, eventually impairing the goal-reaching behavior and increasing the chance of collisions.

It is also worth mentioning that, our framework does not employ the popular neural networks because 
we attempted neural networks and compared with them in our earlier work~\citep{Xu-RSS-20}, and the results revealed that without careful reward engineering, the value function approximated using neural networks cannot guarantee safety.
This observation aligns with similar findings in other recent works utilizing neural networks to impose boundary conditions for solving partial differential equations~\citep{raissi2019physics,sukumar2022exact}.

\subsection{Necessity of Reducing the Mesh-based Dimensions}
\begin{figure*}[t]
   \centering
    \subfloat[]{\label{fig:mc-performance}\includegraphics[height=6.3cm,width=7cm]{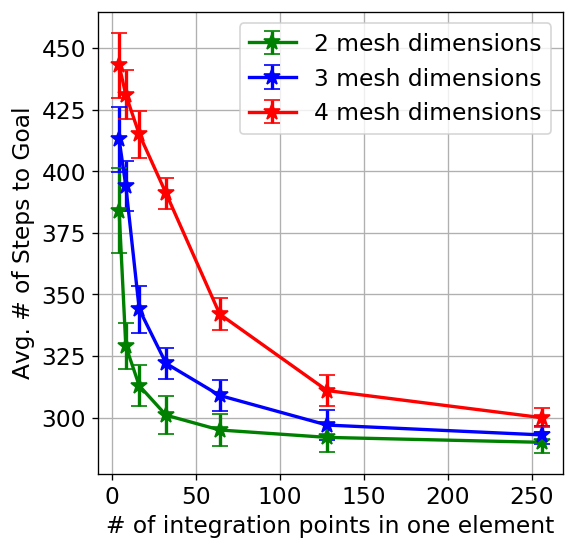}} 
    \hspace{0.4cm}
    \subfloat[]{\label{fig:mc-time}\includegraphics[height=6.3cm,width=7cm]{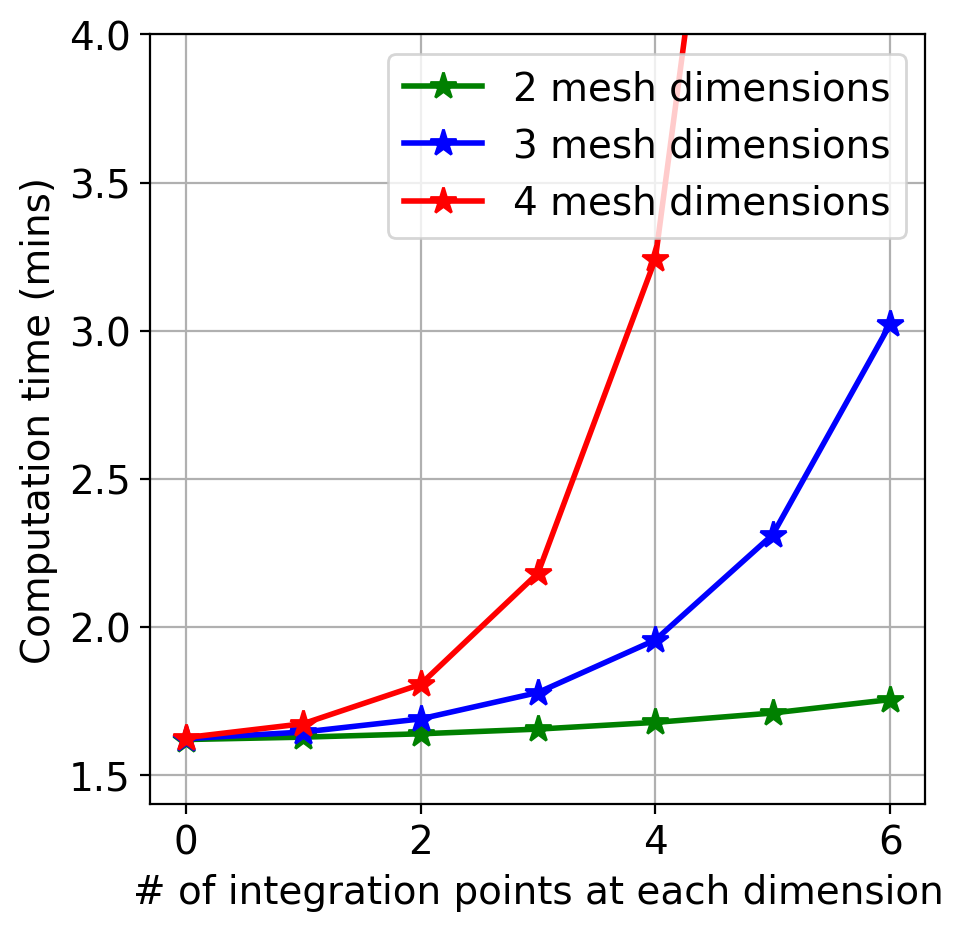}}
    \caption{
    The performance and computation time comparison between the varying mesh dimensions in figures (a) and (b), respectively.
    In (a), the $\mathrm{x}$ axis shows each element's total number of integration points.
    In (b), the $\mathrm{x}$ axis shows the number of integration points at one mesh-based dimension.
    Thus, the total number of integration points in one element is $n^d$, where $n$ is the mesh dimensions, and $d$ is the number of integration points.
    We use the same kernel function as the previous section for the meshless dimensions. 
    All the methods are computed until convergence and tested in $0.25$ ratio obstacle environments.
    }
    \label{fig:monte-carlo-int}
\end{figure*}

\subsubsection{Setup}
\begin{figure*}[t]
    \centering
    \subfloat{\label{fig:}\includegraphics[height=1.4in, width=1.8in]{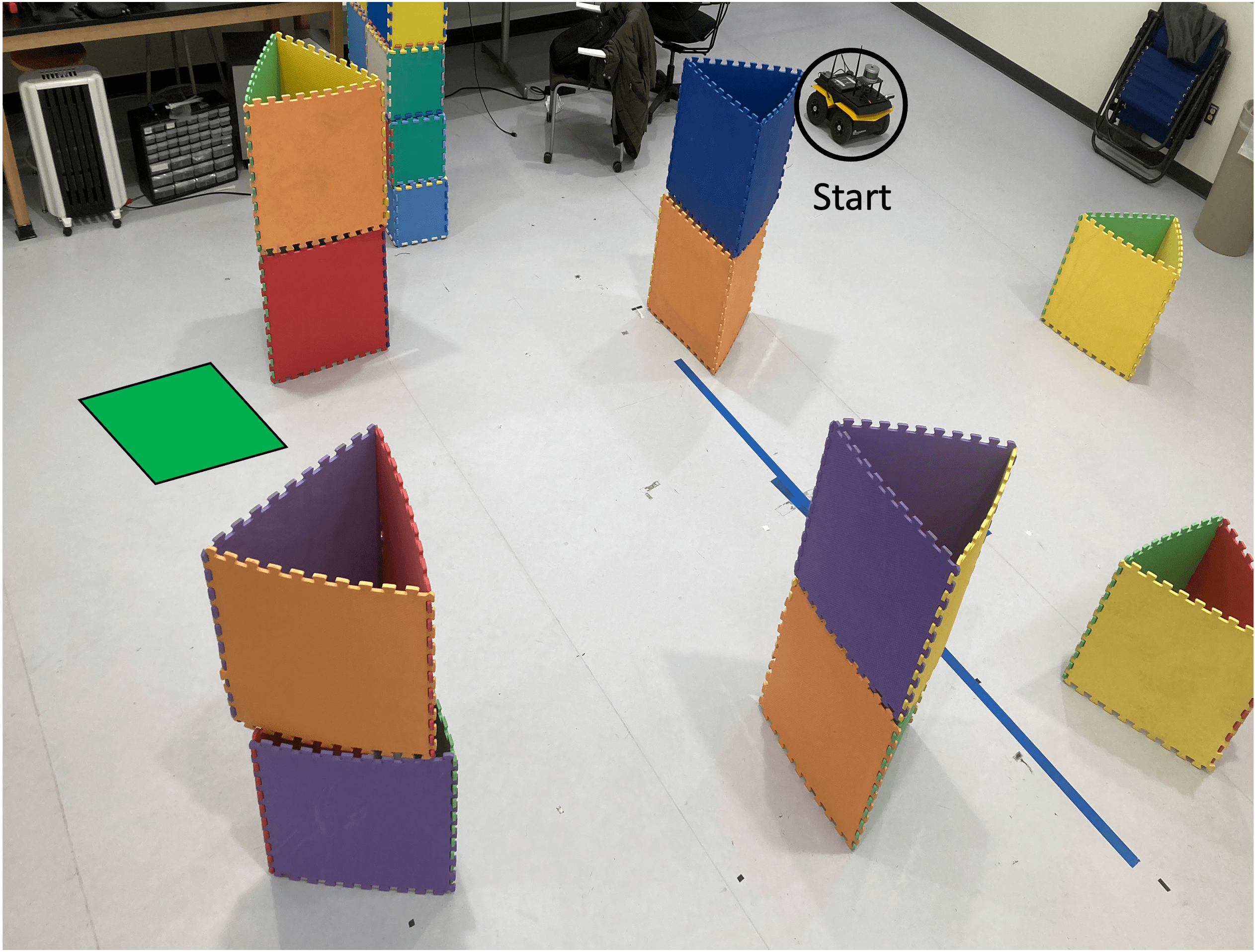}} 
    \subfloat{\label{fig:}\includegraphics[height=1.4in, width=1.6in]{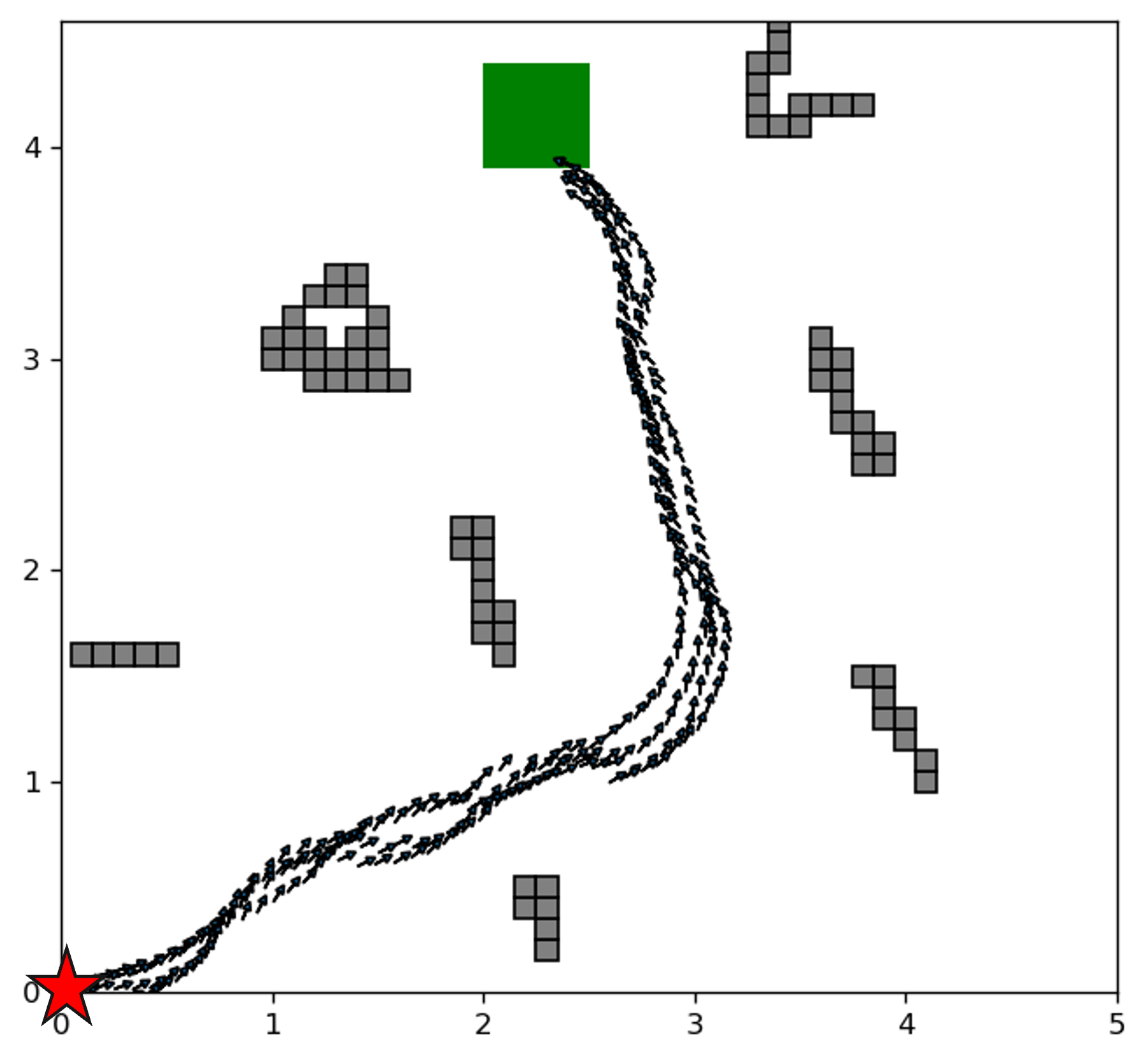}} 
    \subfloat{\label{fig:}\includegraphics[height=1.4in, width=1.6in]{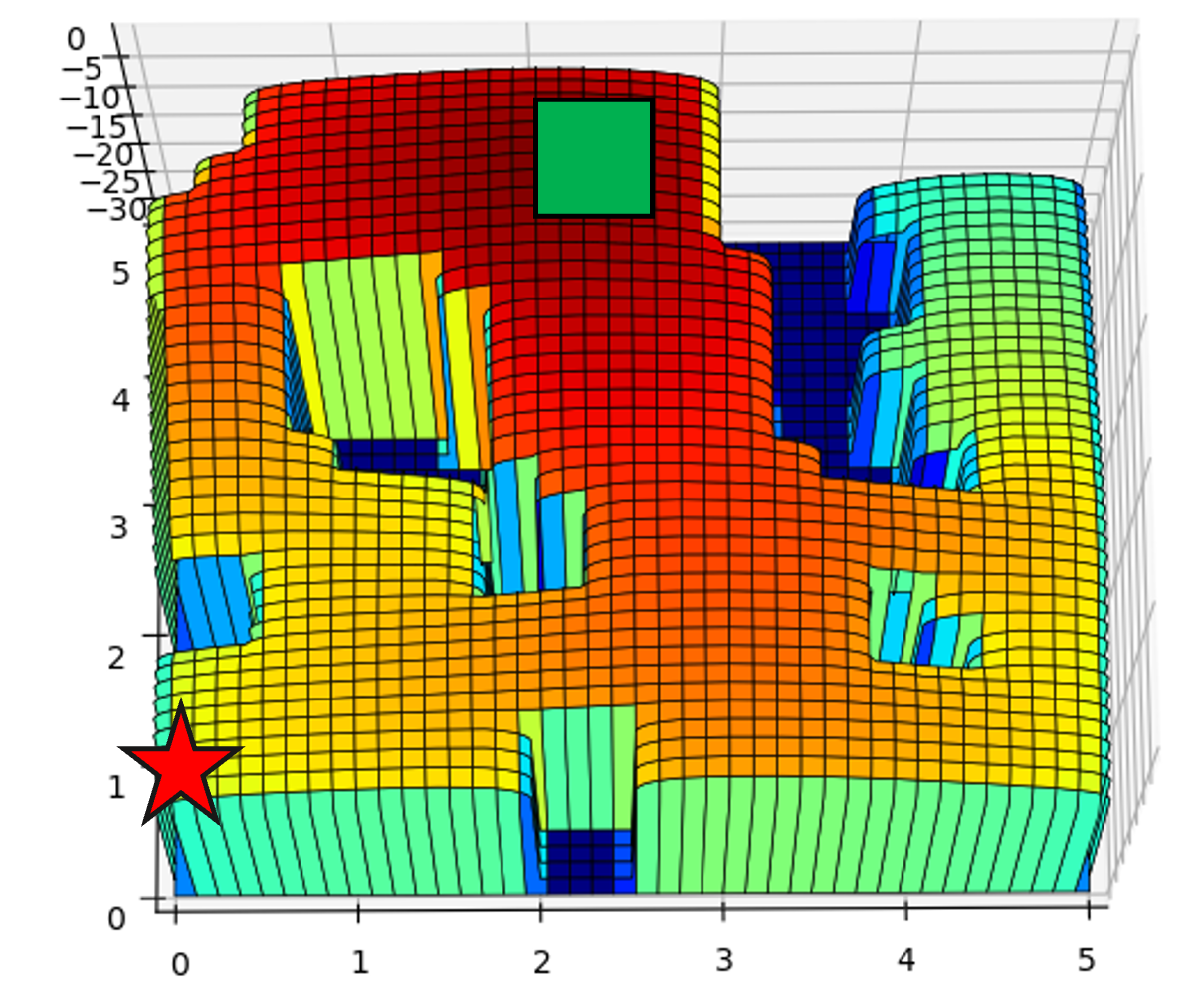}}
    \subfloat{\label{fig:}\includegraphics[height=1.4in, width=1.6in]{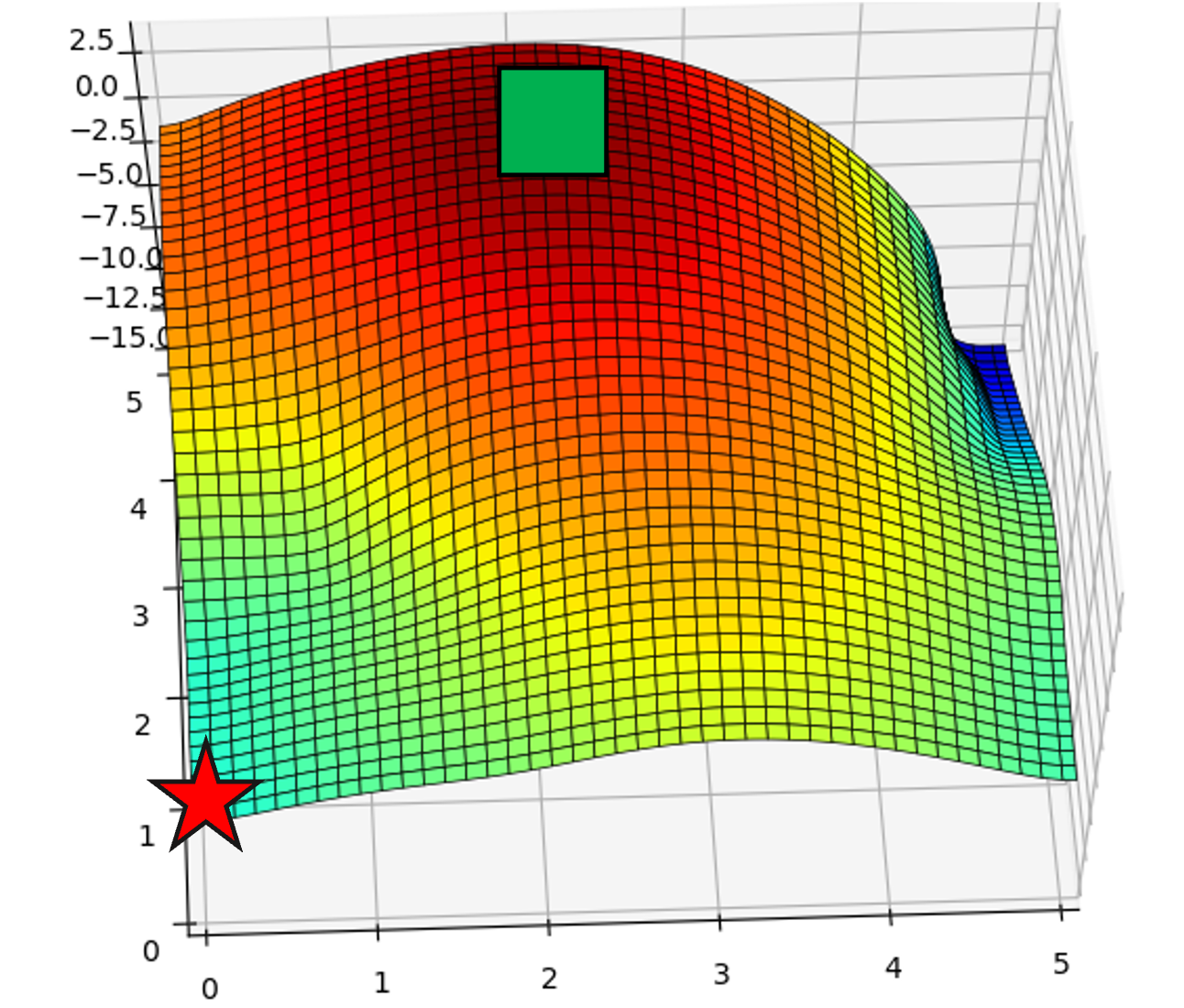}}\\
    \subfloat{\label{fig:}\includegraphics[height=1.4in, width=1.8in]{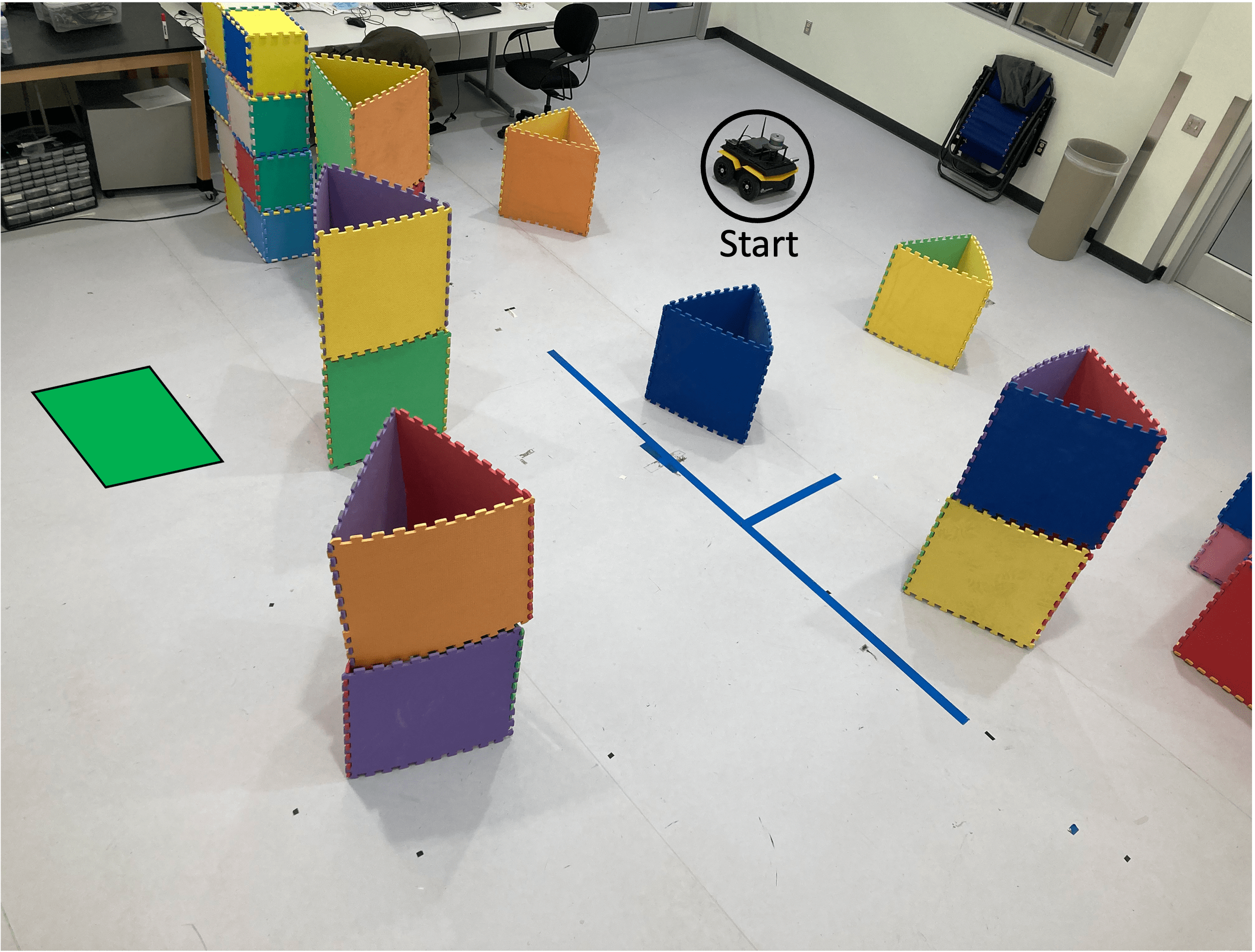}}
    \subfloat{\label{fig:}\includegraphics[height=1.4in, width=1.6in]{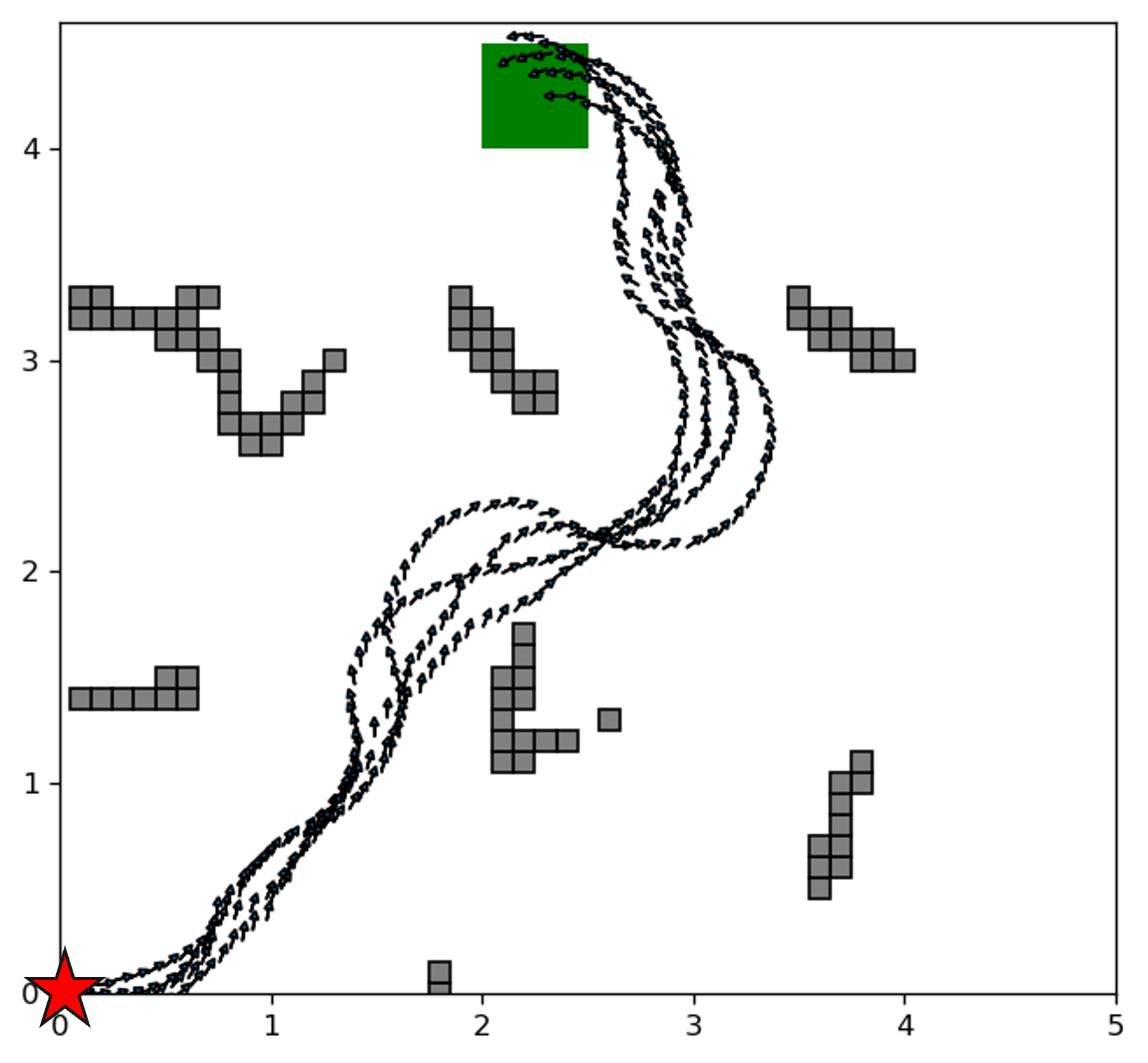}} 
    \subfloat{\label{fig:}\includegraphics[height=1.4in, width=1.6in]{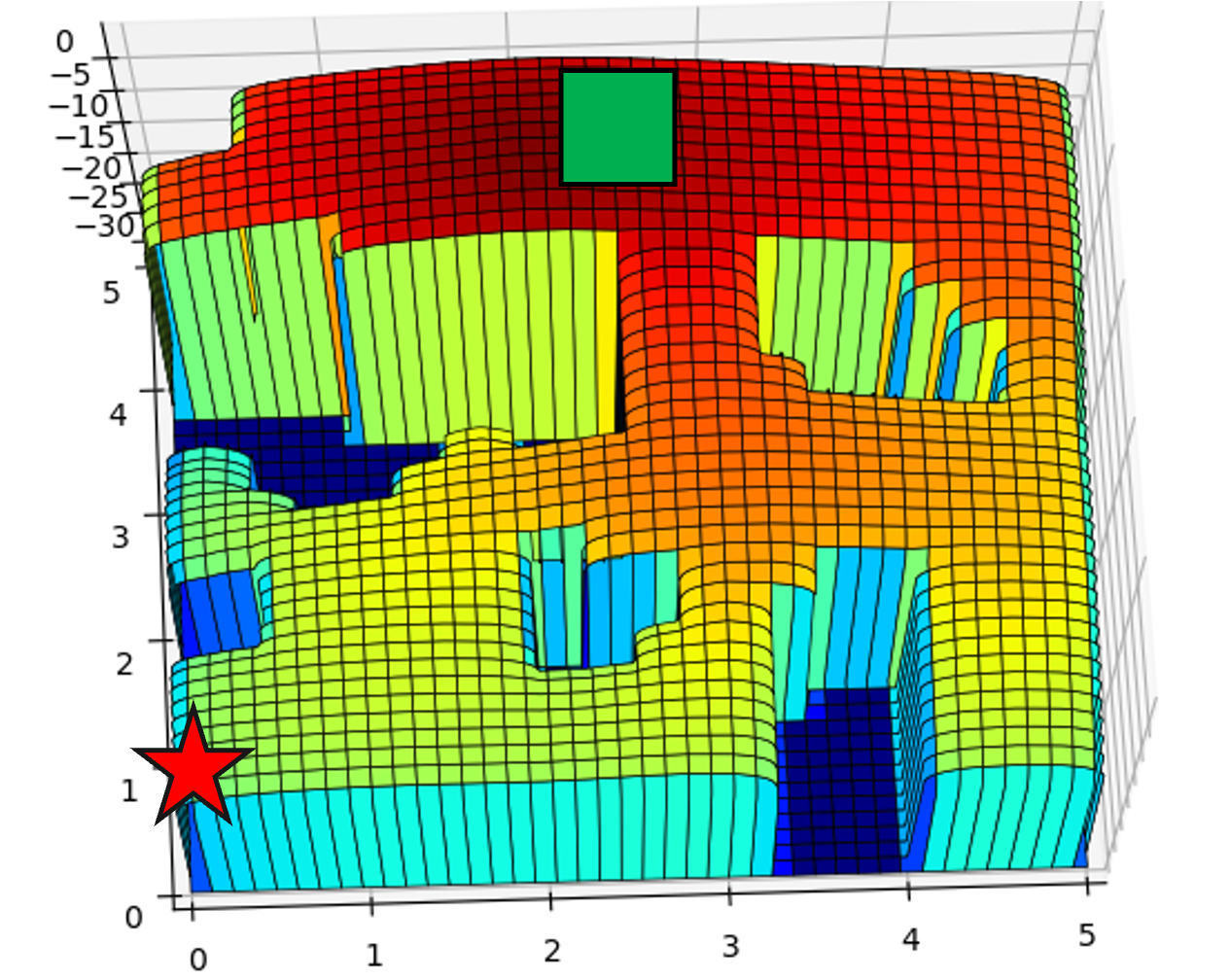}}
    \subfloat{\label{fig:}\includegraphics[height=1.4in, width=1.6in]{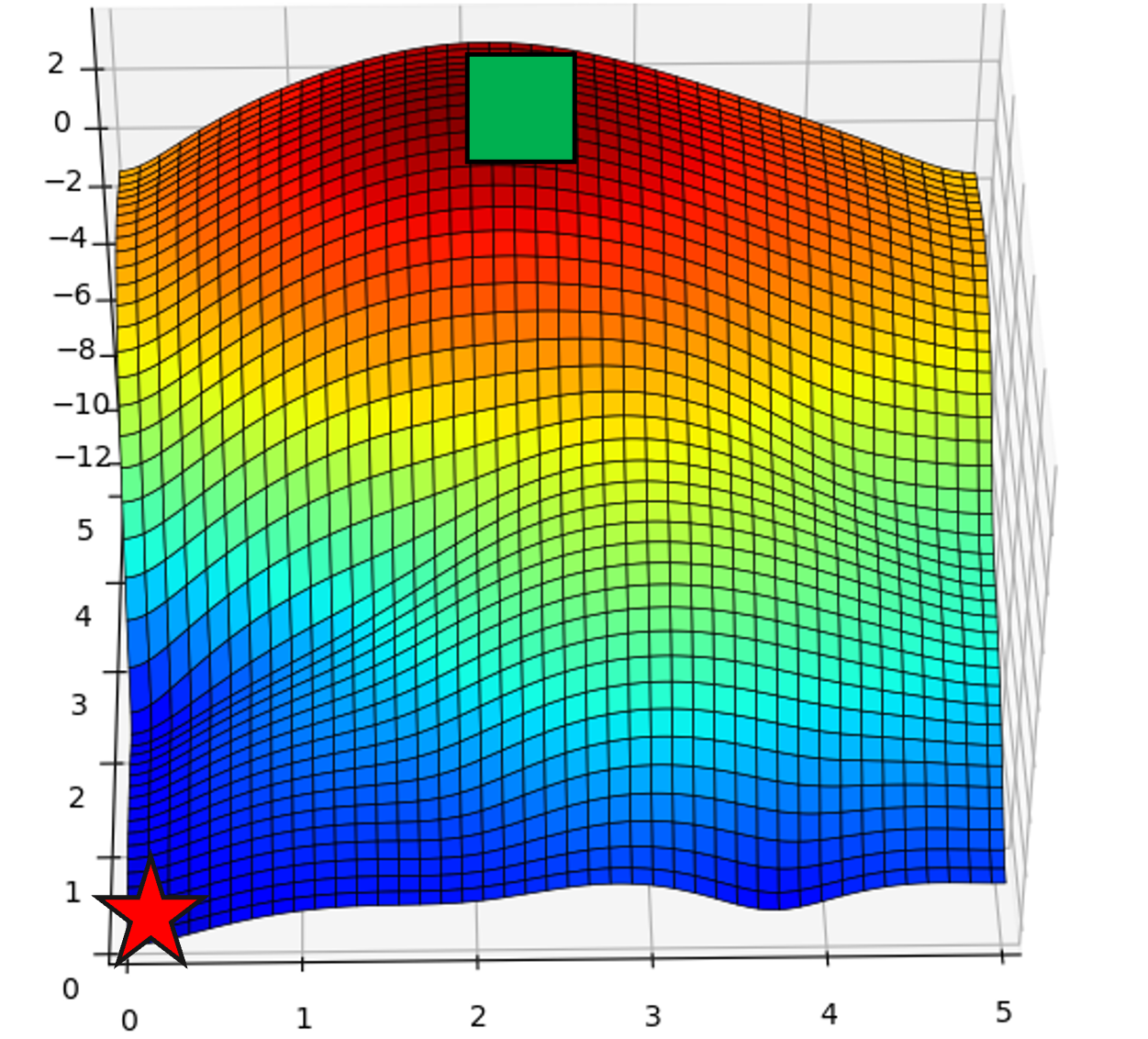}} \\ 
    \subfloat{\label{fig:}\includegraphics[height=1.4in, width=1.8in]{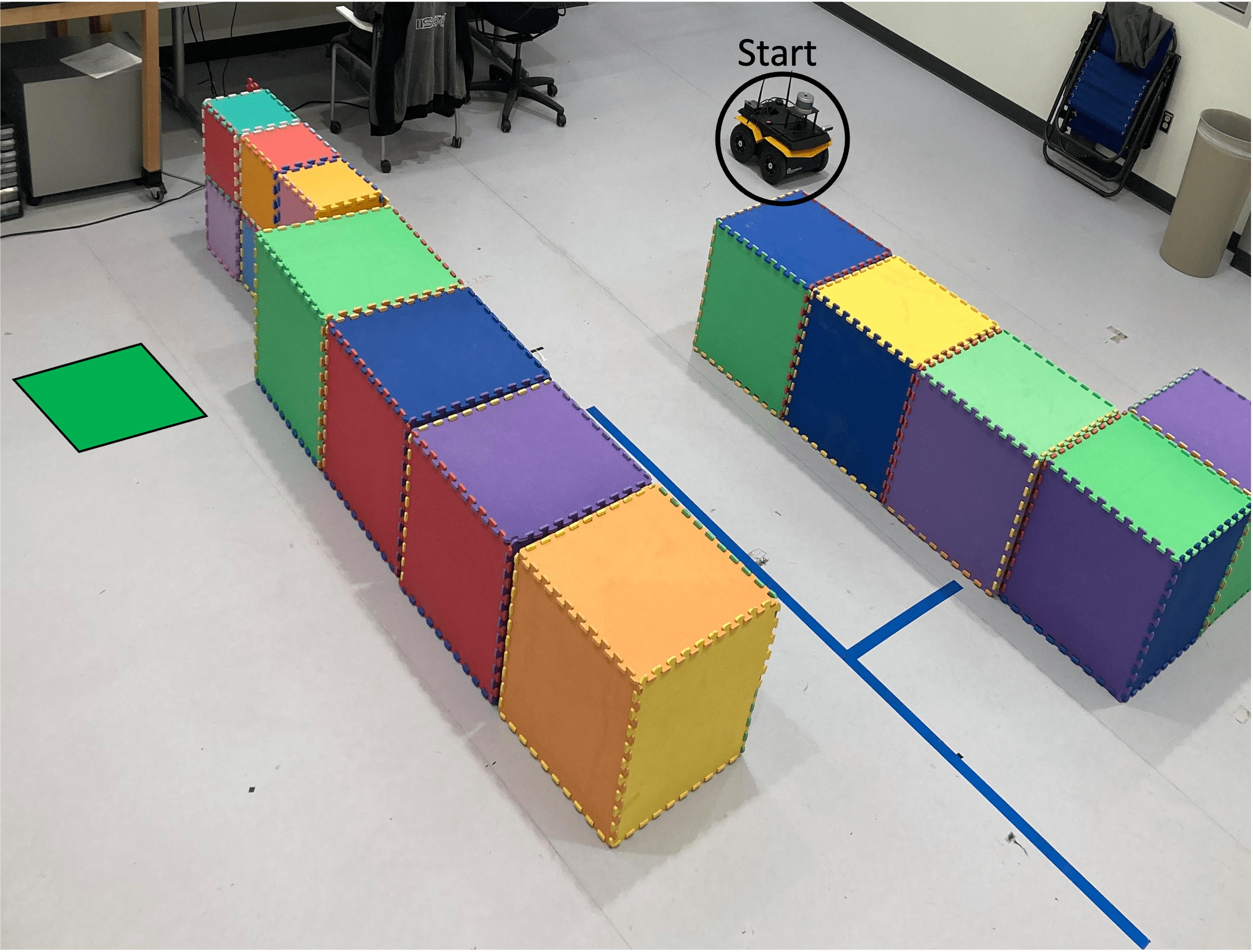}} 
    \subfloat{\label{fig:}\includegraphics[height=1.4in, width=1.6in]{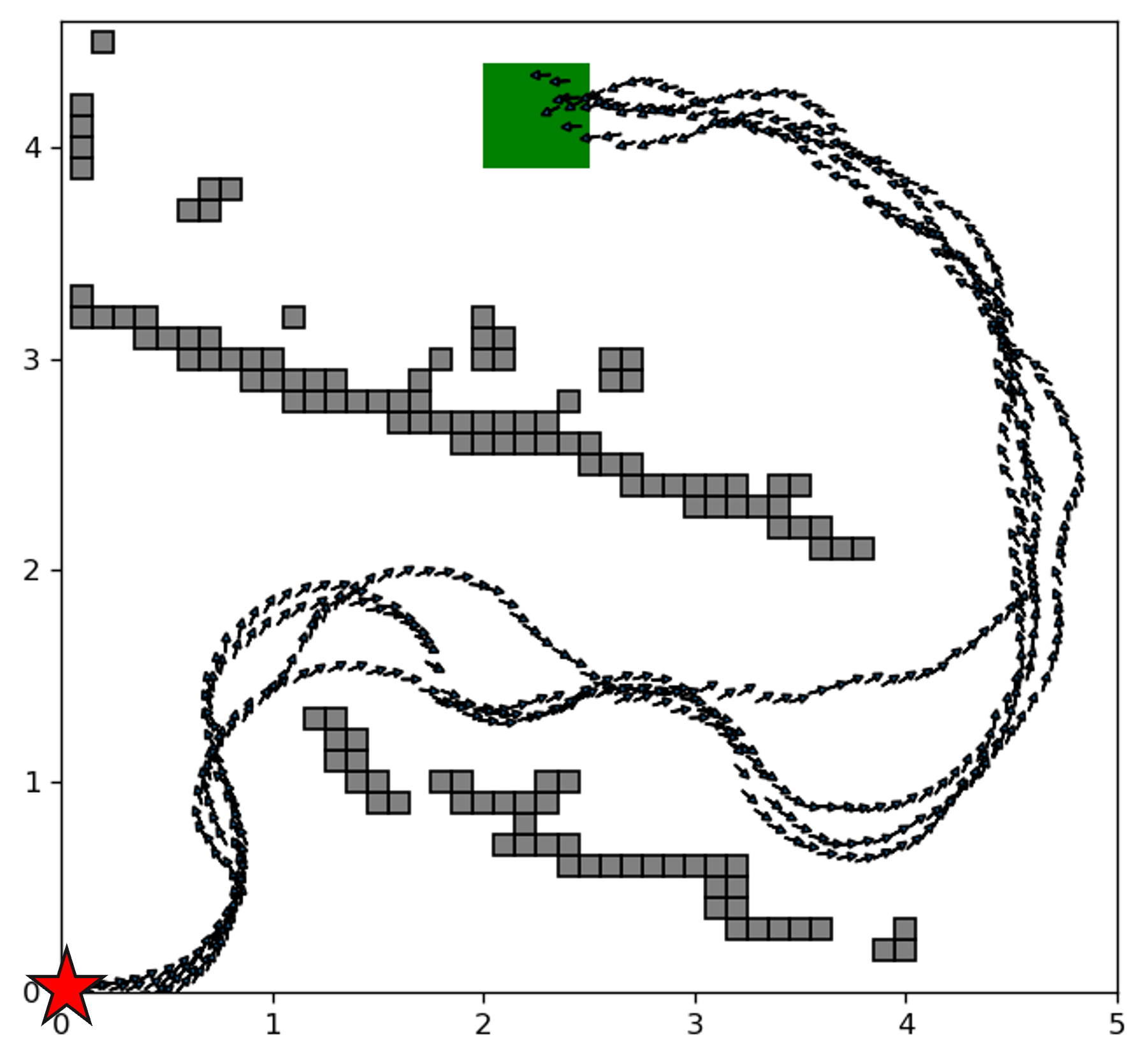}}
    \subfloat{\label{fig:}\includegraphics[height=1.4in, width=1.6in]{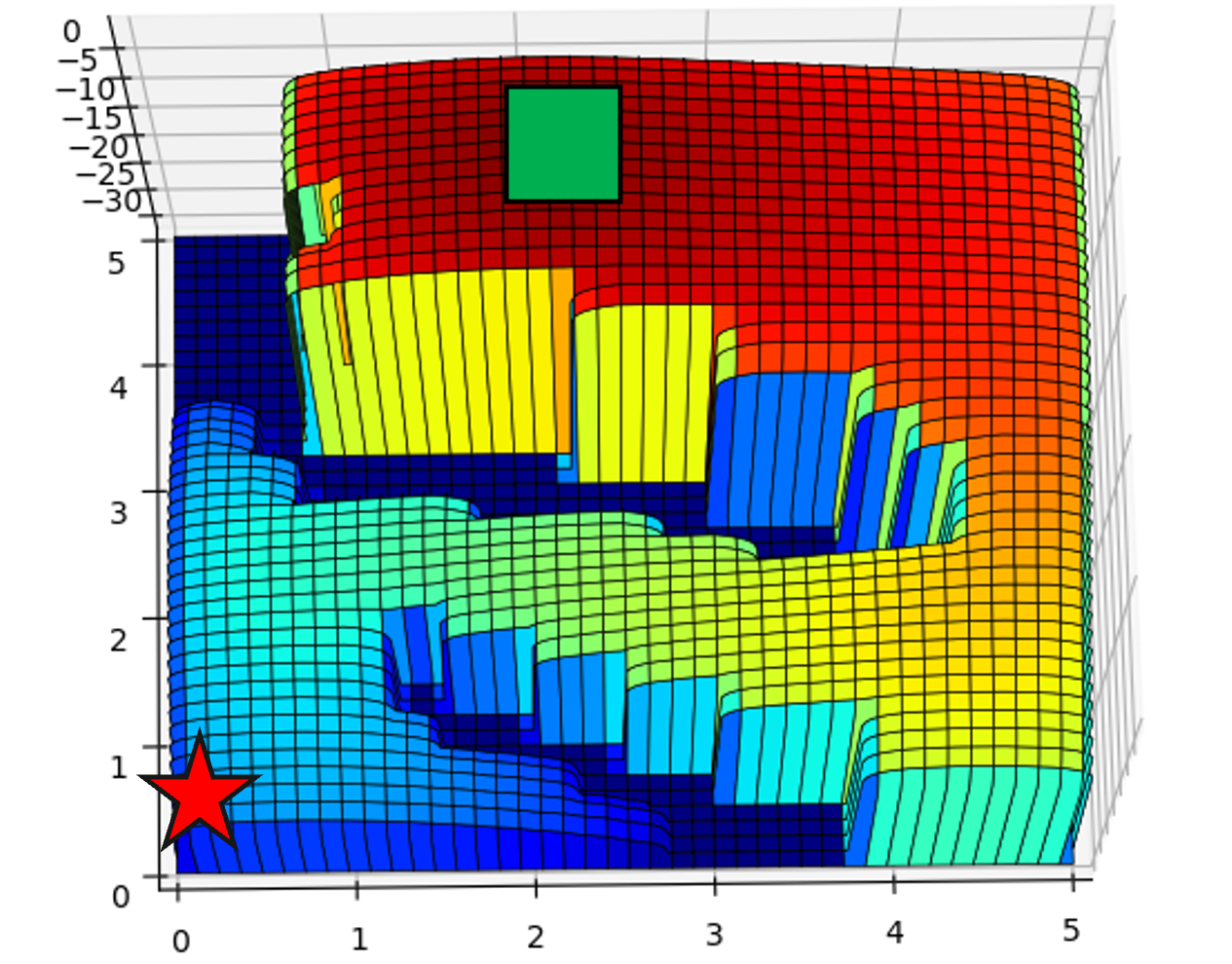}}
    \subfloat{\label{fig:}\includegraphics[height=1.4in, width=1.6in]{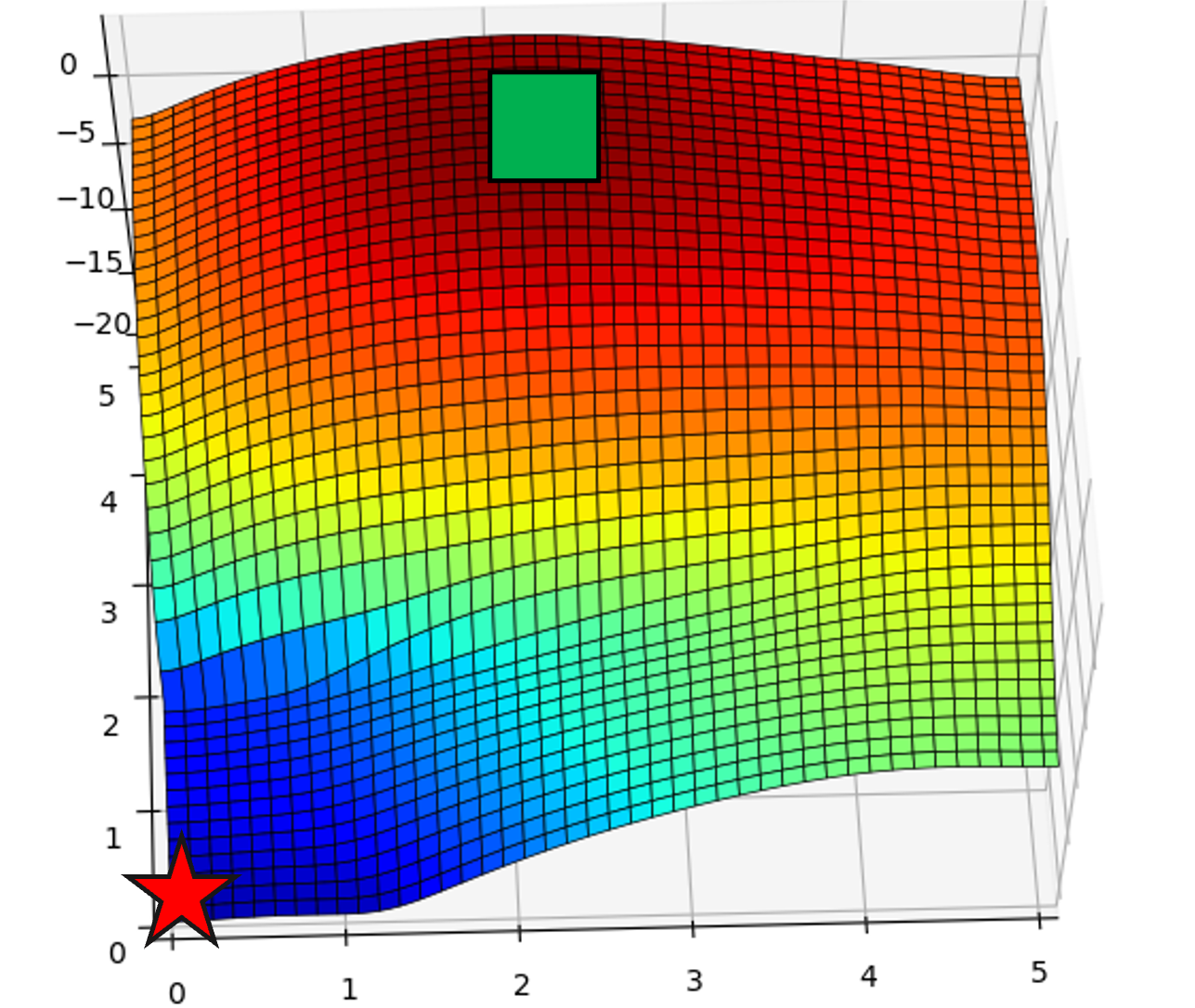}}
    \caption{
    Each row shows the navigation result of one obstacle configuration.
    The columns from left to right represent the environment, its corresponding obstacle map and multi-trial trajectories, our method's value function, and the kernelized value function, respectively.
    In all the figures, the green squares denote the goal locations.
    The first column shows snapshots of obstacle configurations where the robot is placed at the starting position; The robot starting positions are highlighted as red stars  in both the obstacle maps and the visualizations of the value function in the subsequent columns.  (Note, the viewing perspective in the first column differs from other columns due to spatial constraints that restricted our options for capturing angles.)
    The value functions in the last two columns are shown in log scale for the $(\mathrm{x},\mathrm{y})$ dimensions at $\theta = \frac{\pi}{2}$.
    The grey squares in the second column represent the obstacles and the goal, and the arrows represent the trajectories of $4$ runs. 
    We ignore the trajectories of the kernelized value function since none of the runs arrived at the goal.
    The noisy trajectory distributions are due to navigating on the smooth floor at a relatively high speed.
    This also implies that since this drift effect is hard to model, it is essential to consider uncertainty during planning.  
    }
    \label{fig:real-world}
\end{figure*}

In the preceding experiment, we observed improved performance compared to baseline methods regarding safety and goal-reaching behaviors. 
In this experiment, we aim to examine the necessity of utilizing the hybrid (mesh-based and meshless) representation when not all state dimensions require strict satisfaction of safety and goal-reaching constraints.
To this end, we conduct two experiments with relatively higher dimensional state spaces.
The first one uses the same simulated environment as described in the previous section but modifies Dubin's car model to include an additional dimension
$\mathrm{x}_{t+1}=\mathrm{x}_t+v_{t}dt\cos{\theta_t}$, 
$\mathrm{y}_{t+1}=\mathrm{y}_t+v_{t}dt\sin{\theta_t}$,
$\theta_{t+1} = \theta_{t} + (\omega+\epsilon)dt$, 
$v_{t+1} = v_t + (a_{v}+\epsilon)dt$,
where $\epsilon$ is the additive control noise, $a=(a_v, \omega)$ is the action representing linear acceleration and angular velocity.
The velocity dimensions are bounded by $v \in [-1, 1]$m/s and $\omega \in [-\pi, \pi]$ rad/s.

In the second experiment, we further demonstrate that with high-level guidance computed by a motion planner (or designated guiding action sequences), we can leverage the flexibility of the kernel representation to scale up to a much higher state space than the grid-based method.
The action consists of linear and angular jerk $a = (\dddot{x}, \dddot{\theta})$. 
This corresponds to a {7}-dimensional state space, whose elements consist of second-order physical quantities, 
$s = (\mathrm{x},\mathrm{y},v,a_v,\theta, \dot{\theta}, \ddot{\theta})$.
The high-level guidance can be in the form of a desired velocity and acceleration profile, facilitating sampling of a set of sparse state points proximal to the provided profile.
These state points can then be used to construct the kernel functions at the non-critical dimensions.
To generate high-level guidance, we address a two-point boundary-value problem.
This involves determining a trajectory from an initial state to a terminal condition at the goal position while ignoring the obstacles. 
We use a sampling-based trajectory optimizer~\citep{theodorou2010generalized,mohamed2023towards} to solve the boundary-value problem and assume the dynamics is a triple integrator. 
This method provides a systematic approach to exploring trajectory space, balancing the demands of efficiency and precision in high-dimensional state spaces.

To make our experiment challenging, the dimension of the environment is limited to be ${10}{m}\times {10}{m}$ with randomly placed squared obstacles each of which has a size of ${1}{m}\times {1}{m}$.
To test the robustness of the methods while keeping the total travel time similar, we fix the initial position but randomly initialize the robot's initial heading, velocity, and acceleration. 
We test the methods in environments with {0}, {5}, and {10} obstacles, which correspond to obstacle occupancy area $A_{obs}$ as ${0}{m^2}$, ${5}{m^2}$, and ${10}{m^2}$, respectively.

\subsubsection{Results}
\begin{table*}[!htbp]
    \centering
    \resizebox{\textwidth}{!}{%

    \begin{tabular}{@{}lccccccc@{}} 
        \toprule
        Method & \multicolumn{3}{c}{Steps to Goal} & \multicolumn{3}{c}{Success Rate} & {Computational Time (s)} \\
        \cmidrule(lr){2-4} \cmidrule(lr){5-7} 
        & {$A_{obs}={0}{m^2}$} & {$A_{obs}={5}{m^2}$} & {$A_{obs}={10}{m^2}$} & {$A_{obs}={0}{m^2}$} & {$A_{obs}={5}{m^2}$} & {$A_{obs}={10}{m^2}$} & \\
        \midrule
        {7}D Mesh: $10^2 \times 3^5$ & 366 & {N/A} & {N/A} & 0.64 & 0 & 0 & 151 \\
        {7}D Mesh: $10^2 \times 4^5$ & 349 & {N/A} & {N/A} & 0.70 & 0 & 0 & 428 \\
        {7}D Mesh: $10^2 \times 5^5$ & 325 & 385 & 391 & 0.75 & 0.49 & 0.12 & 1921 \\
        \addlinespace 
        {2}D Mesh + {5}D Kernel: $10^2 \times 200$ & 361 & 396 & N/A & 0.72 & 0.17 & 0 & 144 \\
        {2}D Mesh + {5}D Kernel: $10^2 \times 400$ & 342 & 395 & N/A & 0.89 & 0.34 & 0 & 233 \\
        {2}D Mesh + {5}D Kernel: $10^2 \times 800$ & 317 & 357 & 402 & 0.91 & 0.42 & 0.09 & 407 \\
        \bottomrule
    \end{tabular}}

    \caption{Comparison of the proposed tensor product method with the FEM in {7}-dimensional state space. Experiments were conducted on an 8-core Intel i7 CPU and 16GB RAM PC.
    }
    \label{tab:7state-car}
\end{table*}

In the first experiment, to demonstrate the advantage of the proposed tensor product method, we thoroughly vary the number of integration points in each dimension and evaluate the corresponding effect. 
For the mesh-based dimensions, numerical integration is needed to compute the integral within one element.
In this experiment, we use Monte Carlo integration~\citep{burden2015numerical}, where a representative sample of points is drawn in each element, and the numerical integration is computed based on the samples.
The results, as shown in Fig.~\ref{fig:monte-carlo-int}\subref{fig:mc-performance}, reveal a correlation between the number of integration points and the performance of the proposed method.
An increase in the number of integration points results in a decrease in the number of actions required to reach the goal, indicating improved performance. 
However, as the number of mesh dimensions increases, the number of integration points needed to achieve a similar performance as the lower mesh dimensional variant increases.
This result is unsurprising as a higher-dimensional mesh typically requires more points to be filled to achieve a more accurate integration.
On the contrary, since the integration at the meshless dimension can be carried out efficiently, mentioned in Section~\ref{tensor-product-basis}, the integration procedure in the meshless dimension does not exhibit significant scalability issues as demonstrated in Fig.~\ref{fig:monte-carlo-int}\subref{fig:mc-time}.
This can be observed by the exponential increase in the computational time when the number of dimensions is increased in the mesh.

Table~\ref{tab:7state-car} summarizes the results of the second experiment.
The first three rows display the results using pure mesh methods with different numbers of bins.
For example, the first row with $10^2\times 3^5$ indicates $10$ bins for the first two dimensions and $3$ bins for the rest of the five dimensions. 
The last three rows show the results of tensor product method.
For instance, $10^2\times 800$ in the last row means $10$ bins for the first two dimensions and $800$ kernel method's supporting states in the rest of the dimensions.
These states are sampled around the calculated guiding trajectory using a normal distribution, where the mean is aligned with the values of the guiding trajectory at each timestep, and a uniform variance of {1} is maintained across each dimension. 

Compared to the pure mesh-based method, our method can leverage the high-level guidance in the kernel dimension and thus requires a smaller number of states and reduced computational time to achieve a similar performance. 
In most cases, the two measures, average (time) Steps to Goal and Success Rate, are comparable or even better than the mesh-based methods.
Note that this environmental setup is very challenging, as shown in the Success Rate columns, even though there is no obstacle (with $A_{obs}=0$), the mesh-based methods can achieve a success rate of less than 0.75. This is because the environment is confined,  our motion model is stochastic, and the randomly generated initial state can be difficult for achieving the task (e.g., challenging initial orientation, large initial linear and/or angular velocities). Also, higher-order derivatives of state variables can be more sensitive to noise, with errors easily amplified.

The instances of unsuccessful runs are primarily attributable to the inability of the value function to propagate back to the initial state. 
This issue is similar to the one demonstrated in Appendix~\ref{appendix:value-function}.
In general, it is caused by the sparse-reward setup - the robot only receives a reward of {1} when it arrives at the goal and {0} otherwise.
Compared with the ``dense" reward, the sparse-reward setup needs less reward engineering.
However, with this setup, the value function cannot be properly ``propagated" from the goal to the entire state space via the Bellman operator when we have insufficient discretization of the mesh.
Thus, a reasonable mesh discretization and kernel's support state sampling (possibly non-uniform) dependent on the environment obstacle configuration and the robot's probabilistic transition function are needed to overcome this problem.
The sparse-reward problem is also a major challenge when using deep neural networks and sampling-based reinforcement learning algorithms (e.g., actor-critic methods). 
Without enough state coverage, the agent's behavior may become sub-optimal~\citep{devidze2022exploration}.

We also observe that when obstacles become denser the mesh-based methods can be better.
It implies that the hyperparameters for kernel functions in our methods need to be improved (e.g., learned) for high dimensional planning and control problems, leaving room to enhance our methods in future work.

While the success rates of the applied methods do not reach $100\%$, it is worth mentioning that in all our experiments, these methods consistently avoid obstacle collisions with a 0 collision rate, which is the safety property ensured and inherited by the mesh-based method.

\section{Hardware Experiment}
\begin{figure*}[t]
    \centering
    \subfloat[]{\label{fig:}\includegraphics[width=0.31\linewidth]{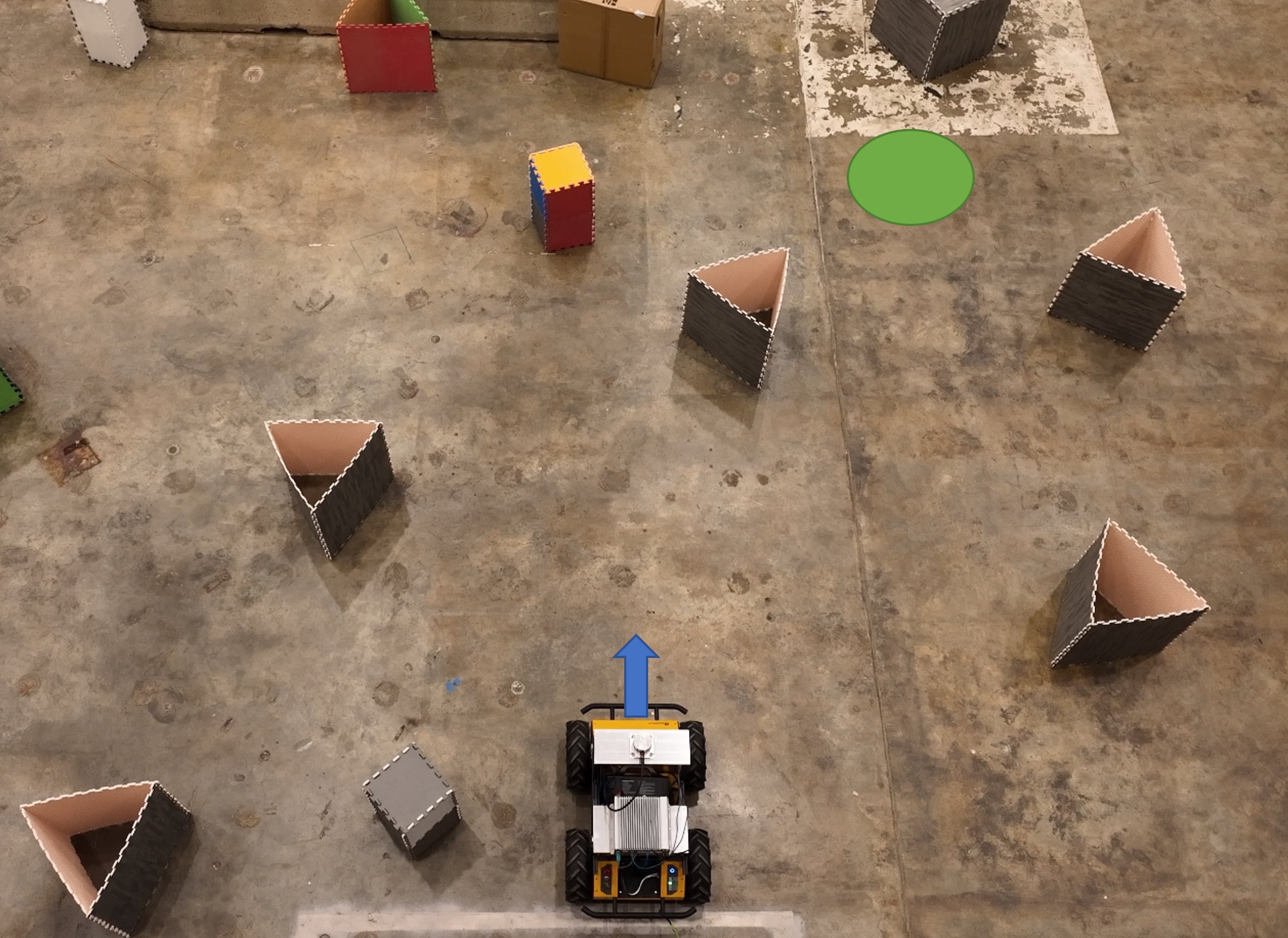}} 
    \hspace{0.1cm}
   \subfloat[]{\label{fig:}\includegraphics[width=0.31\linewidth]{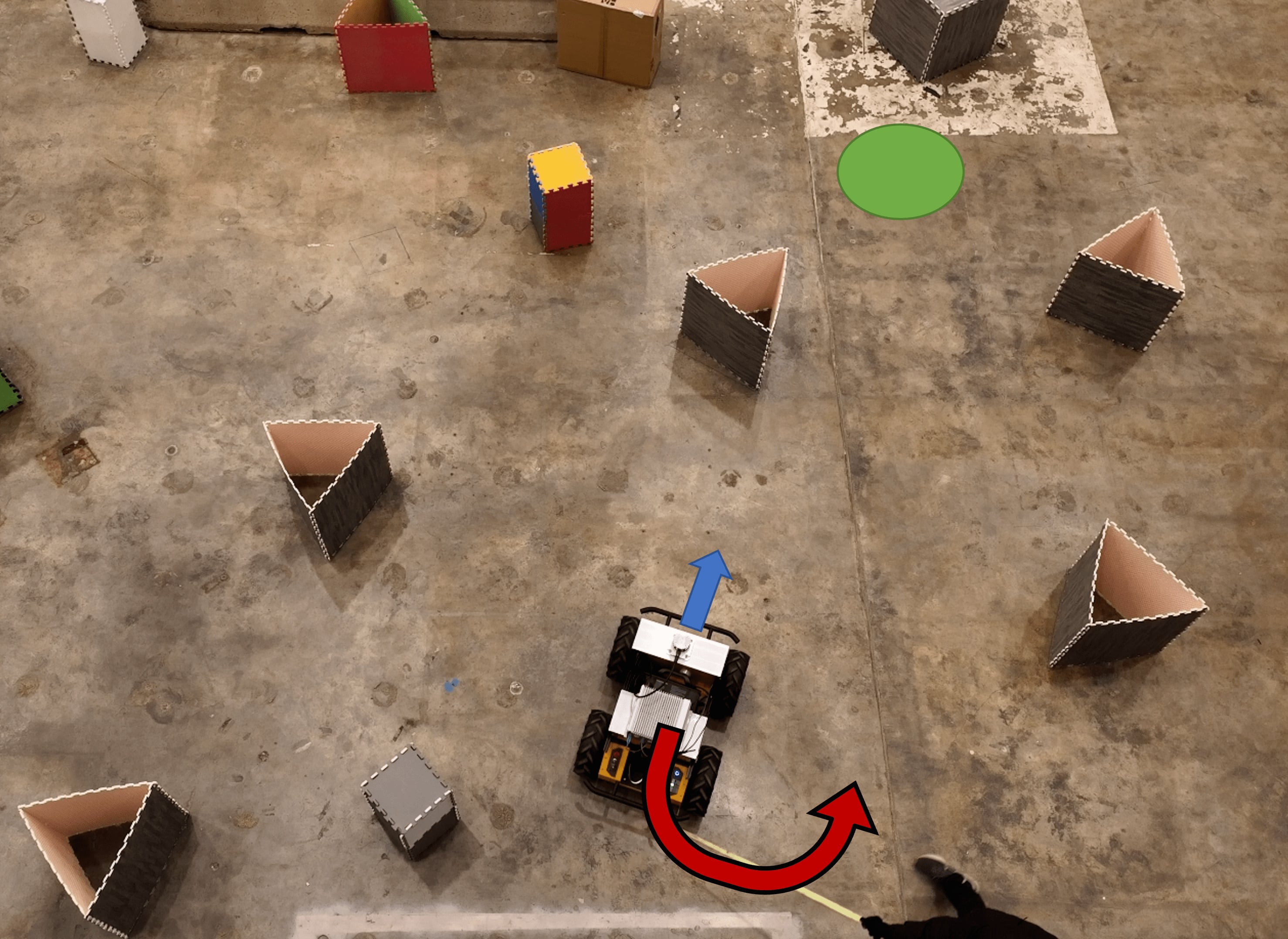}} 
    \hspace{0.1cm}
    \subfloat[]{\label{fig:}\includegraphics[width=0.31\linewidth]{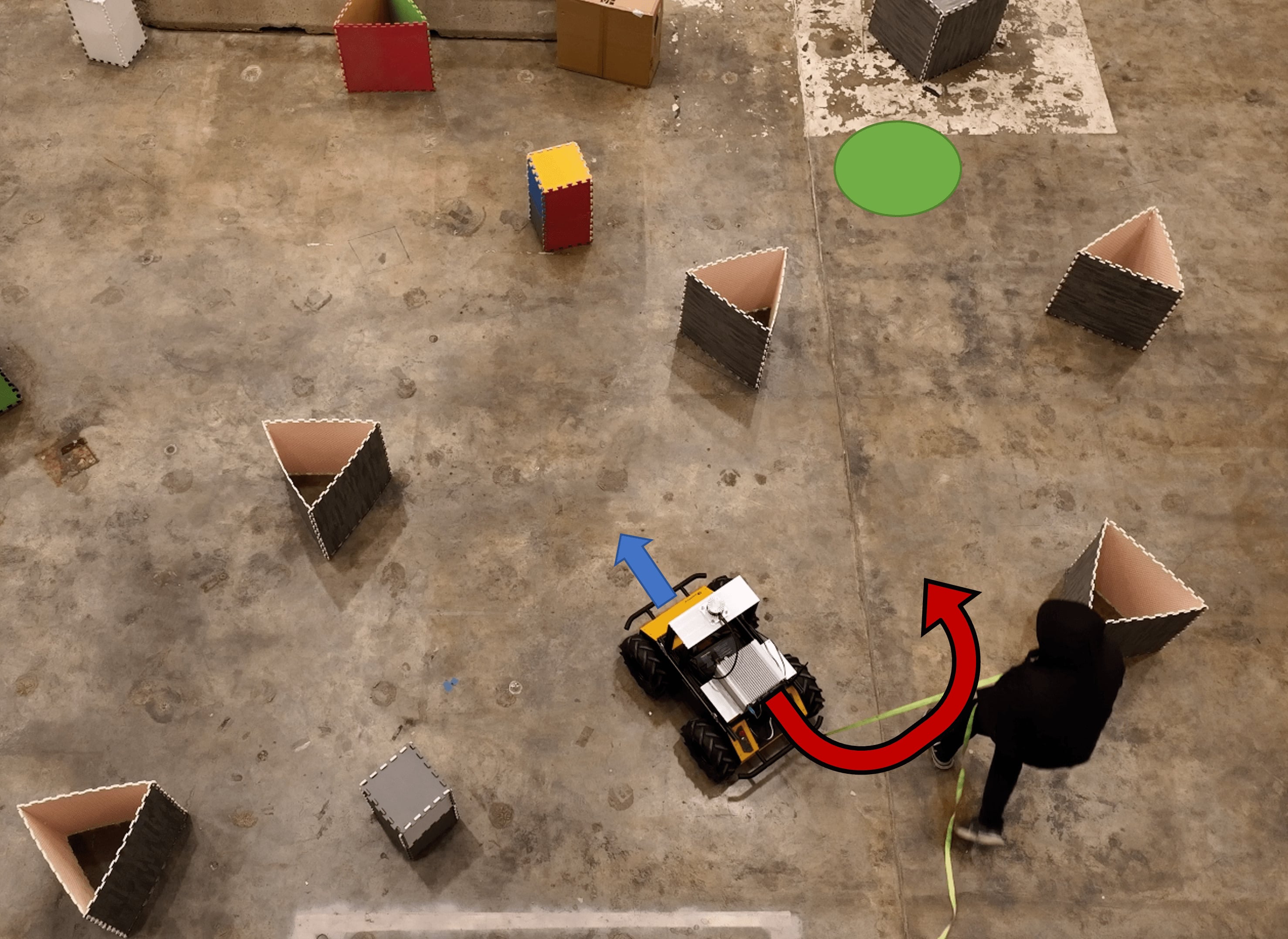}} \\
    \subfloat[]{\label{fig:}\includegraphics[width=0.31\linewidth]{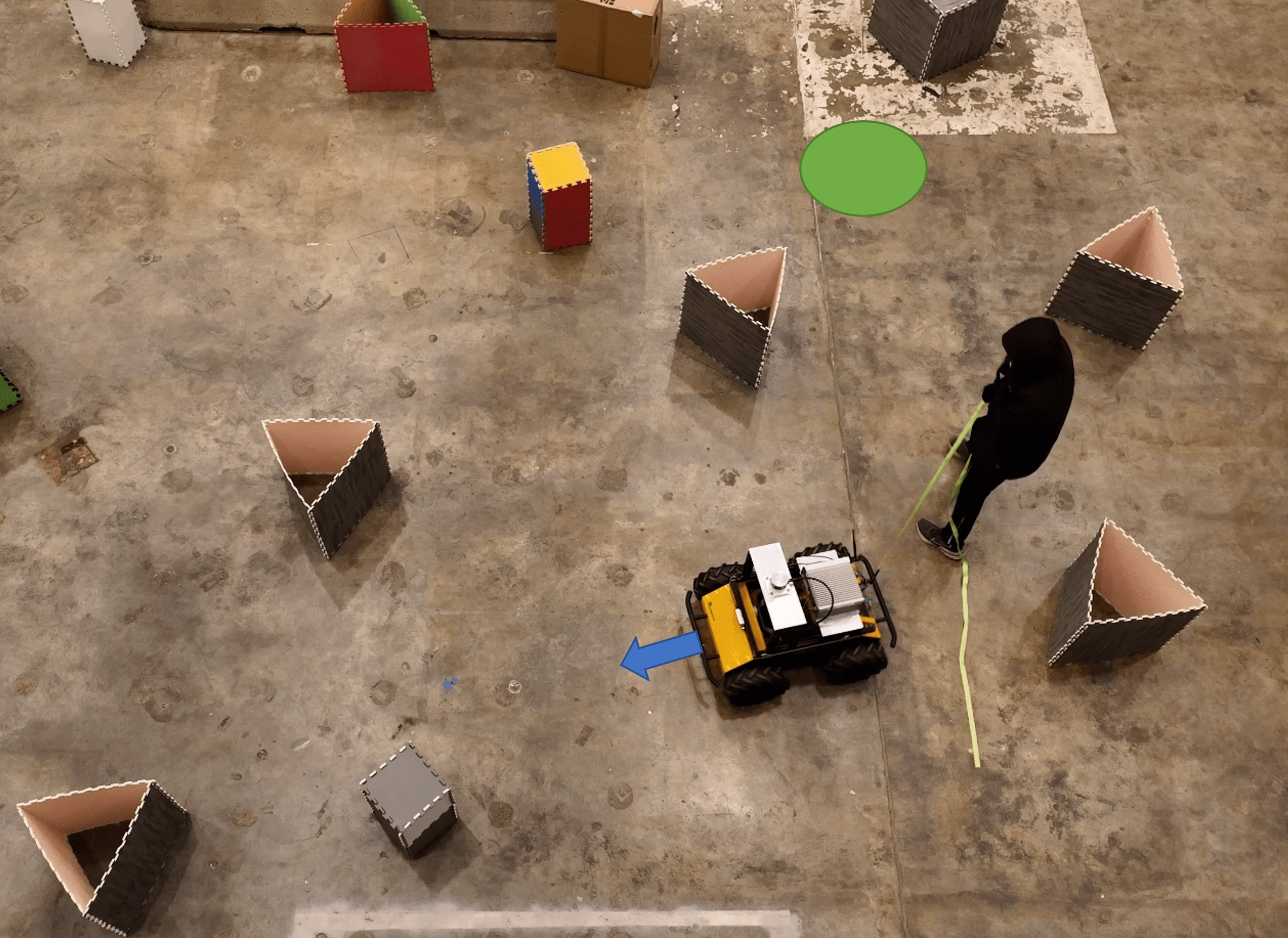}}
    \hspace{0.1cm}
    \subfloat[]{\label{fig:}\includegraphics[width=0.31\linewidth]{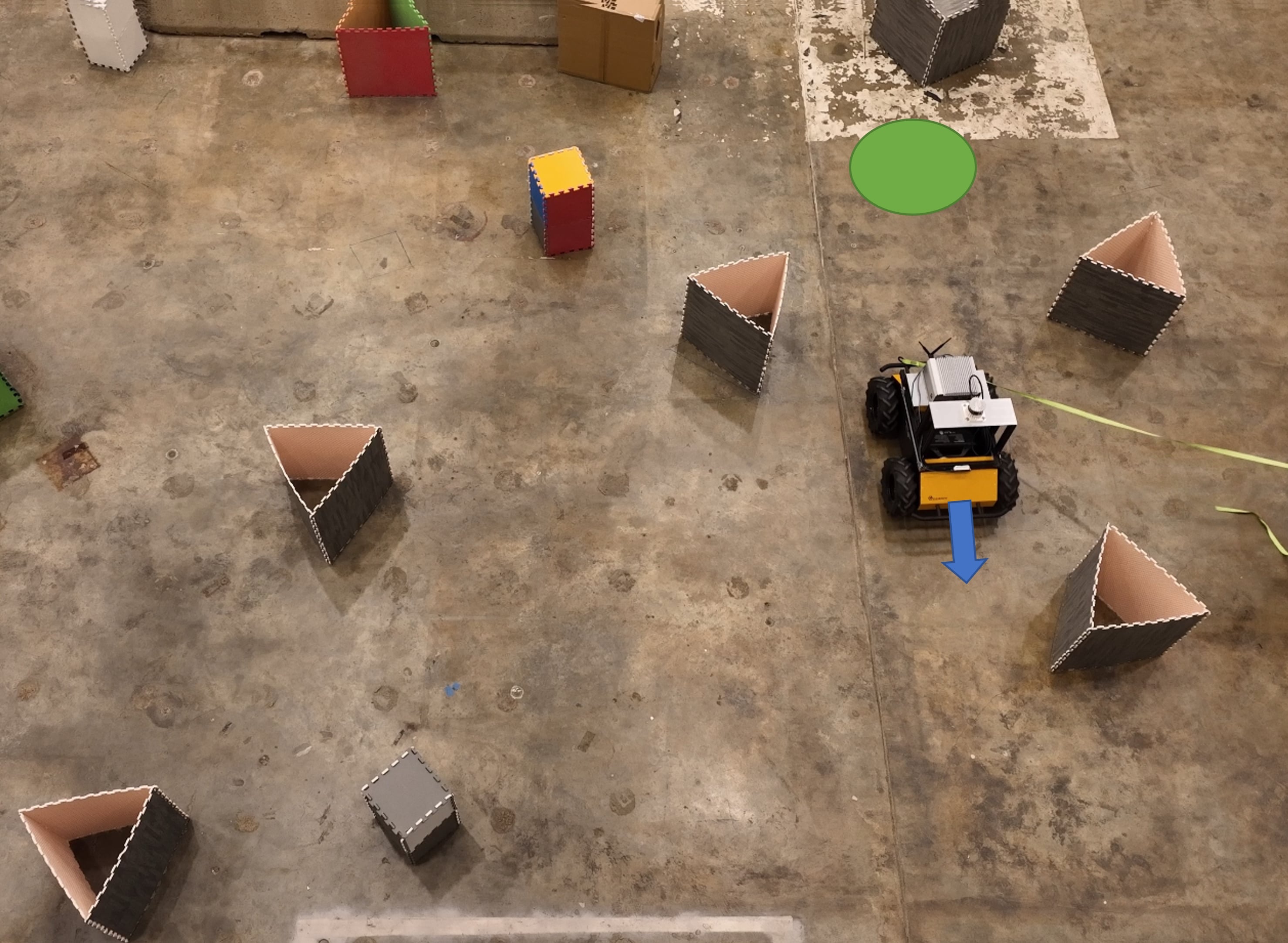}}
    \hspace{0.1cm}
    \subfloat[]{\label{fig:}\includegraphics[height=4.1cm ,width=5.5cm]{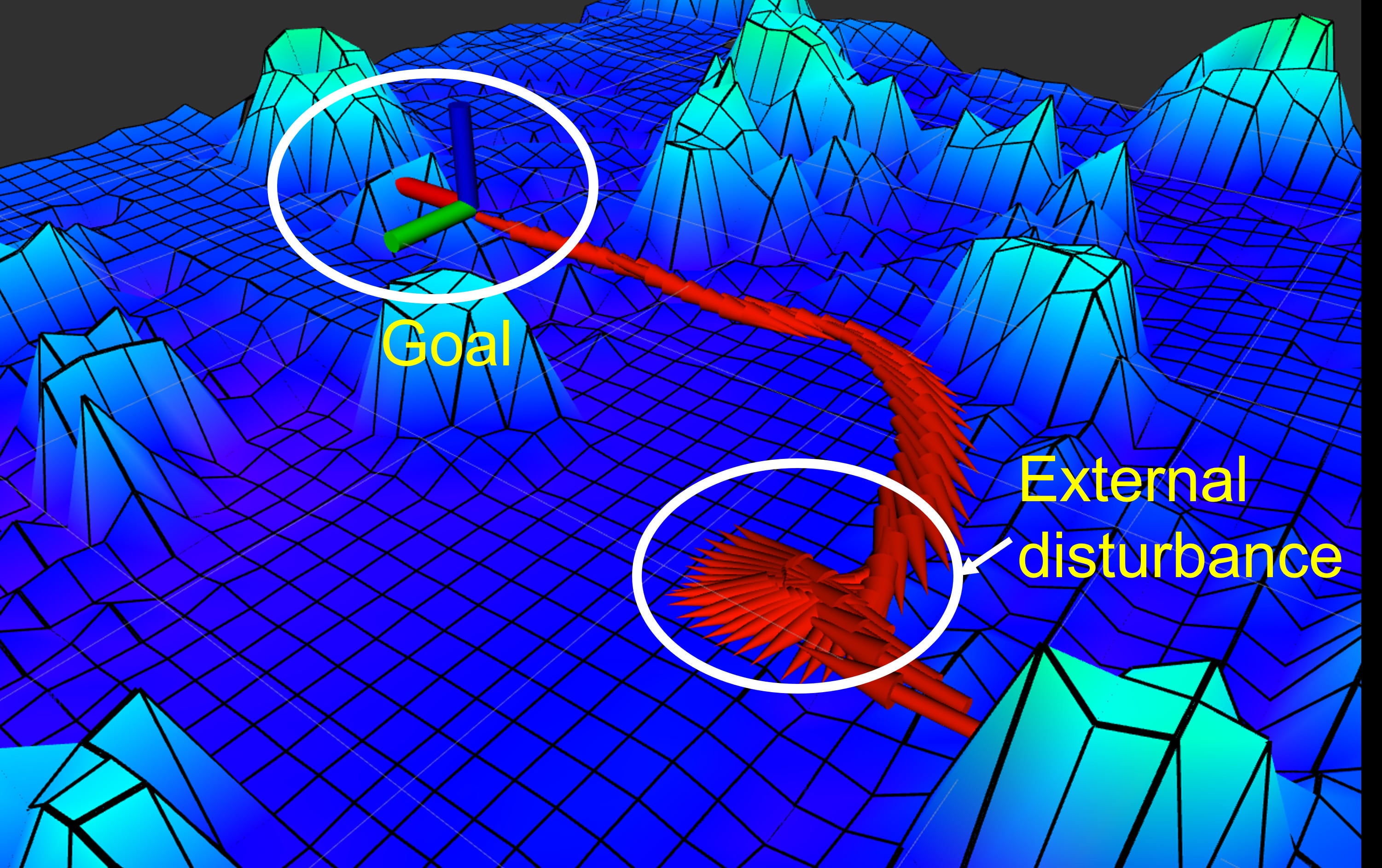}} 
    \caption{
    (a-e) Snapshots of navigation under external disturbances.  The designated goal is shown in the green circular region. 
In the early stages of navigation, depicted in (a) and (b), the robot moves towards the goal. 
However, the occurrence of external disturbances as a result of human interventions, as indicated by red arrows in (b) and (c), led to a significant deviation from the intended trajectory, resulting in a $180$ degree rotation of the robot's heading angle, represented by a blue arrow.
Despite these hindrances, the robot demonstrates its adaptive capabilities by efficiently readjusting its trajectory and successfully reaching the designated goal.
    (f) A snapshot of visualization of the constructed obstacles and the robot trajectory. 
}
    \label{fig:disturbance} \vspace{-10pt}
\end{figure*}

\subsection{Navigation on Slippery Surface}
The first hardware experiment is conducted using a ClearPath Jackal ground robot in a $5m \times 5m$ confined environment with three obstacle configurations shown in the first column of Fig.~\ref{fig:real-world}. 
The first two environments consist of 7 obstacles which are placed approximately $1.2m$ away from each other on average. 
The third environment consists of a $1.2m$ wide and narrow passage, requiring the robot to turn $180$ degree twice to reach the goal.
The robot's maximum speed is set to $1m/s$, relatively fast in a $5m  \times 5m$ confined space with obstacles.
Additionally, due to the smooth floor, this speed can cause noticeable drift which is the disturbance (uncertainty) exerted onto the action outcome. 
In all the environments, the goal is placed at $(2.5m, 4m)$ with respect to the robot's starting position.

We conducted $4$ tests for each obstacle configuration, and the trajectory results are provided in the second column of Fig.~\ref{fig:real-world}. 
{Our method directly computes control signals to the robot instead of using combined global planner, trajectory optimization, and low-level controller to track the trajectories} as in many conventional robotic systems. 
From a robotic navigation system perspective, our method can find the solution from the global path planning to the control problem using a single method. 
The advantage of this integration is that there is no need to engineer heuristic solutions to solve the path deviation problem caused by external disturbance or control errors during the execution. 
As long as the robot is initialized in the planning region, our policy can execute controls to navigate the robot to the goal region without colliding with obstacles.
From a theoretical perspective, the ability to solve this integrated planning and control problem is due to the power of the proposed mechanism, which can satisfy the boundary conditions in the continuous state space. 
This behavior can also be explained by the value function generated using our method in the third column of Fig.~\ref{fig:real-world}, where the value function increases toward the goal and decreases at the boundary of the map and obstacles. 
Similar to the result of the simulated experiment, the kernelized value function blurs the obstacle  and goal values, and the resulting policy either collides with the obstacles or is trapped in a local region.

\vspace{20pt}
\subsection{Navigation Under External Disturbance}
We also conducted a real-world experiment on the ClearPath Husky ground vehicle using the same state space and planner parameters as the previous experiment.

The experiment was conducted in a highbay area. Similar to the previous testing, we also placed random obstacles to challenge our robot. 
In order to test our method under significant motion uncertainty, we generated the external disturbances by intentionally pulling the vehicle in random directions with large forces. 
Snapshots of one trial run are shown in Fig.~\ref{fig:disturbance}.

Specifically, after the vehicle started executing its policy, it experienced periodic jerk motions intervened by humans, causing notable departures from its current states and prompting the robot's heading angle to make abrupt rotations.
Despite these large and abrupt disturbances, 
the policy ensured that the vehicle remained within the state space and maintained safe motions toward the goal while avoiding obstacles, as long as the external disturbances did not disturb the vehicle out of the state space. 
The results demonstrate that the policy proves to be robust enough to handle large disturbances and efficiently adjusts its motion to adaptively realign with the goal. 
Demonstrations can be watched in this video\footnote{\url{https://www.youtube.com/watch?v=URD5Z87jdO0}}.  

\section{Discussion and Future Work}\label{sec:future}

While the preceding experiments highlighted the advantages of our proposed method, its current form exhibits certain limitations, pointing toward avenues for enhancement in our future work.

First, the finite elements approach offers a distinctive advantage in its capability to effectively represent and provide high-degree approximations for diverse geometrical and irregular obstacle shapes~\citep{hughes2012finite}, even though it may result in an increased preprocessing time for element construction.
As our first attempt, the proposed framework is currently implemented by utilizing a {\em basic uniform meshing} strategy, which is an important reason for its slow computation. 
In our future work, we aim to enhance the efficiency of triangular element generation for lower-dimensional states, moving beyond the uniform grid meshes utilized in the experiments.  
This expansion will not only optimize computational resources but also foster improved adaptability. To achieve this, we need to enable the elements to dynamically adjust to the environment, varying the density and geometrical shapes in different regions or sub-spaces. 
Apart from that, leveraging GPUs' parallel architecture has been proved to boost FEM performance substantially by distributing computations across multiple GPU cores~\citep{fu2014architecting,marinkovic2019survey}. (Yet, achieving GPU-based FEM parallelization requires expertise in numerical methods, GPU programming, and efficient memory management.) 
The successful implementation of these advancements is anticipated to significantly boost computational efficiency.

Second, while kernel functions employed to reduce the numerical integration cost in our proposed methods exhibit a useful reproducing property, proper kernel design/selection with efficient hyperparameter optimization becomes crucial with the increasing state dimension, as evidenced in our experiments. 
In addition, the computational cost to construct and solve the linear system of equations in the proposed framework remains heavy. This is another reason that prevents our method from real-time planning and control when the environment and state dimensions become large. 
In practical applications, forward sampling approaches like Model Predictive Control (MPC) and Model Predictive Path Integral (MPPI)~\citep{bemporad1999robust, theodorou2010generalized} prove suitable for real-time control.  
Our future work will concentrate on integrating these ideas to alleviate computational costs, thereby improving efficiency, particularly in handling high-dimensional cases.

Finally, our framework assumes that
the robot's goal and the obstacle configuration remain fixed. However,
in practical scenarios, the goal and obstacles in a planning task may
dynamically change.
This could necessitate the recomputation of the optimal
policy each time a new goal is set up and/or a new obstacle configuration is updated. 
If one can predict the
changed goals or the moving obstacles, 
an extension of the formulation could involve
incorporating a time dimension and utilizing a sequence
of goals or a trajectory of obstacles as boundary conditions.
Additionally, when the robot operates in a prior unknown environment, it is necessary to re-compute the policy once information about the environment is received.
We can enhance the policy update efficiency by bootstrapping the computation using the previous value function.
This approach is viable because successive frames typically contain overlapping environment information.
Thus, only a portion of the value function requires updating, which reduces computational time.

\section{Conclusion}\label{sec:conclusion}
For a stochastic motion planning and control task for mobile robots, an ideal value function is expected to induce a continuous-state control policy that optimizes performance while accounting for the uncertain outcomes of potential collisions.
To address this challenge, we propose a new method for generating the value function based on the tensor product between the finite element interpolation polynomial and the kernel function.
This approach enables the value function to satisfy safety-critical boundary conditions.  
Through extensive simulated and real-world experiments, our method demonstrates its ability to steer mobile robots away from collisions with obstacles while simultaneously achieving efficient motion behavior.

\section*{Acknowledgement}\label{sec:acknowledgment}

The authors extend their appreciation to Durgakant Pushp for his assistance in conducting the hardware experiment on navigation under external disturbances. Additionally, the authors sincerely thank all anonymous reviewers for their invaluable and constructive feedback, which has significantly contributed to the improvement and enhanced quality of this manuscript. 

\section*{Declaration of conflicting interests}
The authors declared no potential conflicts of interest with respect to the research, authorship, and/or publication of this article.

\section*{Funding} 
The authors disclosed receipt of the following financial support for the research, authorship, and/or publication of this article. 
This work has been supported by the Army Research Office and was accomplished under Cooperative Agreement Number W911NF-22-2-0018: Scalable, Adaptive, and Resilient Autonomy (SARA).  The views and conclusions contained in this document are those of the authors and should not be interpreted as representing the official policies, either expressed or implied, of the Army Research Office or the U.S. Government. The U.S. Government is authorized to reproduce and distribute reprints for Government purposes notwithstanding any copyright notation herein.
This work was also partially supported by NSF: RI: Small: Exploiting Symmetries of Decision Theoretic Planning for Autonomous Vehicles (grant no 2006886), and NSF: CAREER: Autonomous Live Sketching of Dynamic Environments by Exploiting Spatiotemporal Variations (grant no 2047169).


\vspace{20pt}
\appendix
\section*{Appendices}
\addcontentsline{toc}{section}{Appendices}
\renewcommand{\thesubsection}{\Alph{subsection}}

\subsection{Issues of Value Function Propagation}\label{appendix:value-function}
\begin{figure*}[t]
    \centering
    \subfloat[]{\label{fig:app-value-succ}\includegraphics[width=0.4\linewidth]{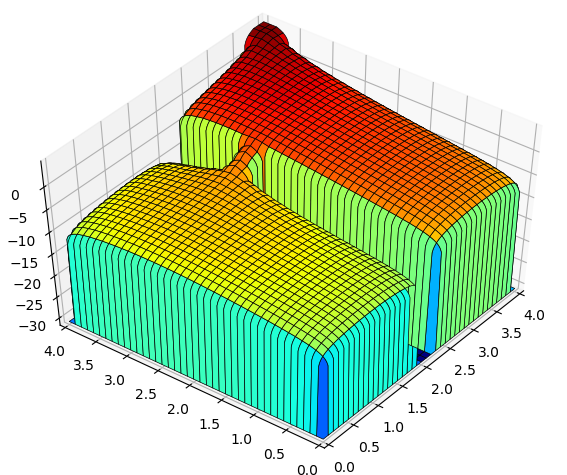}} \quad
    \subfloat[]{\label{fig:app-value-failure}\includegraphics[width=0.4\linewidth]{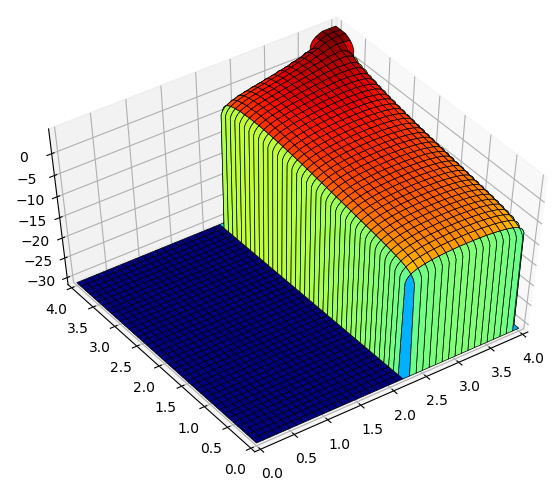}}  \\
    \subfloat[]{\label{fig:appe-traj-succ}\includegraphics[width=0.4\linewidth]{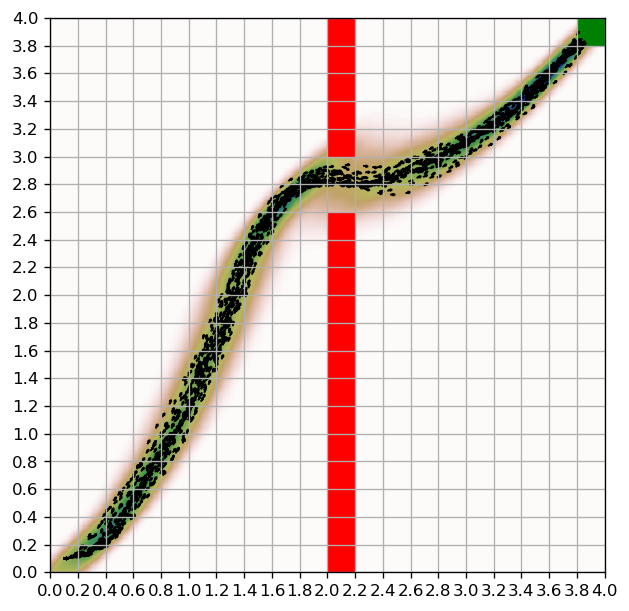}} \quad \quad
    \subfloat[]{\label{fig:app-traj-failure}\includegraphics[width=0.4\linewidth]{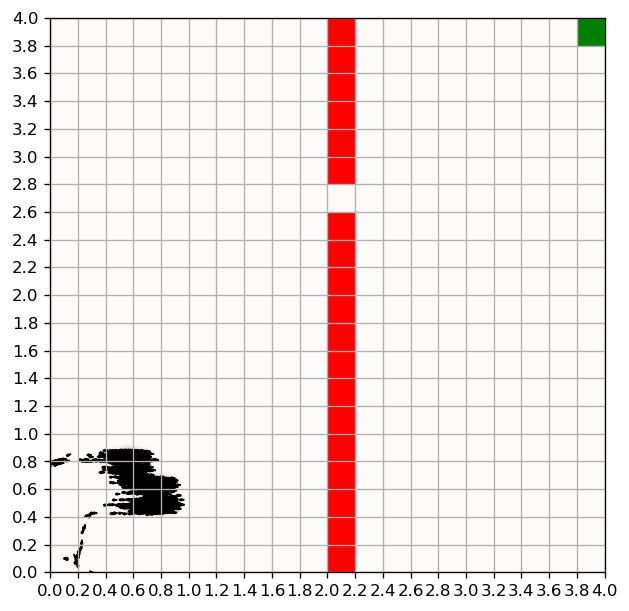}} 
    \caption{
    \small
    (a)(b) are the value functions computed using our method in environments shown in (c)(d).
    In (c), the narrow passage has 0.4m wide, which is equal to the resolution of two square elements.
    In (d), the narrow passage has 0.2m wide, which is equal to the resolution of one square element.
    The value functions are shown in log scale for the $(x,y)$ dimensions at $\theta = \frac{\pi}{2}$.
    }
    \label{fig:appendix-example} \vspace{-10pt}
\end{figure*}
In this section, we discuss the issue of what type of obstacle configurations can prevent the value function from propagating through the state space using an example shown in Fig.~\ref{fig:appendix-example}.
We can observe that the value function in Fig.~\ref{fig:appendix-example}\subref{fig:app-value-failure} cannot propagate through the narrow passage.
Thus, the resulting policy exhibits a random behavior because every state-value at that region is $0$.
The mesh-based method ensures the boundary conditions through the element nodes: the value function coefficients at the obstacle boundary nodes are zero.
Thus, if all the nodes belong to the boundary like the narrow passage element in Fig.~\ref{fig:appendix-example}\subref{fig:app-traj-failure}, that element is treated as a boundary element with $0$ values, which prevents the value function from propagating through the space. 
This issue is not due to the inherit flaws of the method, so it can be solved as shown in Fig.~\ref{fig:appendix-example}\subref{fig:app-value-succ}.
If one more free state is added between the narrow passages, the value can be propagated and the policy can make robot arrive at the goal without collision.
The adaptive discretization can be applied to refine the mesh at these elements, which will be our future work.

\subsection{Construction of Kernel and Polynomial Tensor Product Basis in 3-dimensional State Space}\label{appendix:3d-construction}

Let $(\theta, x, y)^T$ denote the state, where $(x, y)$ is the location, and $\theta$ is the rotation angle. Let $\nabla \triangleq (\partial_{\theta}, \partial_x, \partial_y)^T$. Here the notations denote the partial derivative. For example, $\partial_x v(\theta, x, y)=\frac{\partial}{\partial x}v(\theta, x, y)$.
Consider the following 3-dimensional diffusion equation
\begin{align}\label{apd:diffusion-pde-3D}
&-\nabla\cdot D(\theta, x, y)\nabla v(\theta, x, y) + 
q^T(\theta, x, y)\,\nabla v(\theta, x, y) \nonumber\\
&+ 
b(\theta, x, y)\,v(\theta, x, y) = f(\theta, x, y), 
\end{align}
with boundary conditions
\begin{subequations}\label{boundary-condition-3D}
\begin{eqnarray}
   &{}& D(\theta, x, y)\,\nabla v(\theta, x, y)\cdot \hat{\bm{n}} = 0, \mbox{  on } \partial\Omega_1 \label{boundary-condition-3D-a}\\
    &{}& v(\theta, x, y) = \hat{u}(\theta, x, y), \mbox{on } \partial\Omega_2 \label{boundary-condition-3D-b}
\end{eqnarray}
\end{subequations}
where $\hat{\bm{n}}$ is the unit of outward normal to $\partial \Omega$, and $\hat{u}(\theta, x, y)$ is a given boundary value function.

We construct the approximation solution as the following form
\begin{subequations}\label{kernel-poly-v}
\begin{align}
    v^h(\theta, x, y) &= \sum_{i=1}^{N_1}\sum_{t=1}^{N_2} a_{i,t}\,k(\theta, \xi_t)\phi_i(x, y) \\
    &\triangleq \sum_{i, k} a_{i,t}\, v^h_{i,t}(\theta, x, y),
\end{align}
\end{subequations}
where $\xi_i$ is the supporting states in the dimension of rotation angle. 

The weak formulation of the problem using trial and test functions leads to the linear algebraic equation
\begin{align}
   KA = F,
\end{align}
where $A=(a_{1,1}, ..., a_{1, N_2}, ..., a_{N_1, N_2})^T$ is the vector of unknown parameters, $K$ stiffen matrix, and $F$ a constant vector.

\subsubsection{Construction of Stiffen Matrix $K$}~\\

The matrix $K$ becomes
\begin{align}\label{mixed-stiffen-K}
    K_{(i, t), (j, l)}=&\int_{\Omega^h} (\nabla v^h_{i,t}\cdot D\nabla v^h_{j, l} + v^h_{i,t}\,q^T\,\nabla v^h_{j, l} \nonumber\\
    & + b\,v^h_{i, t}\,v^h_{j, l})\,d\theta dxdy.
\end{align}

Details of computing each component:
\begin{subequations}
\begin{align}\label{derivative-x}
   \nabla v^h_{i,t} &= \nabla k(\theta, \xi_t)\phi_i(x, y) \\
    &=  \left( \partial_{\theta} v^h_{i,t},  \partial_{x}v^h_{i,t}, \partial_{y}v^h_{i,t} \right)^T\\
     &=  \left( \phi_i(x, y)\partial_{\theta} k(\theta, \xi_t),  k(\theta, \xi_t)\partial_{x}\phi_i(x, y), \right.\nonumber\\
     &\left. k(\theta, \xi_t)\partial_{y}\phi_i(x, y) \right)^T.
\end{align}
\end{subequations}
Let $D = \begin{bmatrix}
D_{\theta\theta}, & D_{\theta x}, & D_{\theta y}  \\
D_{x\theta}, & D_{x x}, & D_{x y}  \\
D_{y\theta}, & D_{y x}, & D_{y y}  \\
\end{bmatrix}$, and this matrix is symmetric, for example, $D_{\theta x} = D_{x\theta}$.
Then we have
\begin{align}
    &\nabla v^h_{i,t}\cdot D\nabla v^h_{j,l} \nonumber\\ 
    &= D_{\theta\theta}\partial_{\theta} v^h_{i,t}\partial_{\theta} v^h_{j,l} + D_{\theta x}\partial_{\theta} v^h_{i,t}\partial_{x} v^h_{j,l} + D_{\theta y}\partial_{\theta} v^h_{i,t}\partial_{y} v^h_{j,l} \nonumber\\
    &+ D_{x\theta}\partial_{x} v^h_{i,t}\partial_{\theta} v^h_{j,l} + D_{xx}\partial_{x} v^h_{i,t}\partial_{x} v^h_{j,l} + D_{xy}\partial_{x} v^h_{i,t}\partial_{y} v^h_{j,l}\nonumber\\
    &+ D_{y\theta}\partial_{y} v^h_{i,t}\partial_{\theta} v^h_{j,l} + D_{yx}\partial_{y} v^h_{i,t}\partial_{x} v^h_{j,l} + D_{yy}\partial_{y} v^h_{i,t}\partial_{y} v^h_{j,l}
\end{align}
Each component can be integrated as
\begin{align}
    &\int_{\Omega^h} D_{\theta\theta}\,\partial_{\theta} v^h_{i,t}\partial_{\theta} v^h_{j,l}\,d\theta dxdy \nonumber\\
    &= \int_{\Omega^h}D_{\theta\theta}\,\phi_i(x, y)\phi_j(x, y)\,\partial_{\theta} k(\theta, \xi_t)\partial_{\theta} k(\theta, \xi_l)\,d\theta dxdy \nonumber\\
    &=\int_{\Omega^h_{xy}} \hat{D}_{\theta\theta}\phi_i(x, y)\phi_j(x, y)\,dxdy, 
\end{align}
where 
\begin{align}\label{type-two-derivatives}
   \hat{D}_{\theta\theta} = \int_{\Omega^h_{\theta}}D_{\theta\theta}\,\partial_{\theta} k(\theta, \xi_t)\partial_{\theta} k(\theta, \xi_l)\,d\theta.
\end{align}
For $\hat{D}_{\theta\theta}$, the integral of derivatives of kernels can be pre-computed, or it may be approximated by quadrature rule. 

Assume that we take Gaussian kernel, $k(\theta, \xi_t) = \frac{1}{\sqrt{2\pi}\kappa}\,\exp\left(-\frac{1}{2\kappa^{2}}(\theta-\xi_t)^2\right)$, where $\kappa$ is a scale parameter, then
\begin{align}
   &\hat{D}_{\theta\theta} \nonumber\\
   &= (2\pi\kappa^2)^{-1}\int_{\Omega^h_{\theta}} \kappa^{-2}D_{\theta\theta}(\theta - \xi_t)  \kappa^{-2}(\theta - \xi_l) \nonumber\\
   &\exp\left(-\frac{1}{2\kappa^{2}}\left(2\,( \theta - \frac{\xi_t + \xi_l}{2} )^2 + \frac{1}{2}(\xi_t - \xi_l)^2\right)\right)\,d\theta\nonumber\\
   &=\exp\left(-\frac{1}{4\kappa^{2}}(\xi_t -\xi_l)^2\right)\times(2\pi\kappa^6)^{-1}\times\nonumber\\
   &\int_{\Omega^h_{\theta}} D_{\theta\theta}(\theta - \xi_t)(\theta - \xi_l) \exp\left(-\frac{1}{\kappa^{2}}\,( \theta - \frac{\xi_t + \xi_l}{2} )^2\right)\,d\theta\nonumber\\
   &\approx \frac{1}{8\sqrt{\pi}\kappa^{5}}\,D_{\theta\theta} (\xi_l - \xi_t)^2\exp\left(-\frac{1}{4\kappa^{2}}(\xi_t -\xi_l)^2\right),
\end{align}
where $D_{\theta\theta}\triangleq D_{\theta\theta}(\theta, x, y)$ should be evaluated at $\theta=\frac{\xi_t +\xi_l}{2}$ in the last equation, i.e., $D_{\theta\theta}(\frac{\xi_t +\xi_l}{2}, x, y)$. Note the last step evaluation can be made in various approximate approaches. What we put here is the a simple approximation. 

The integral of $D_{\theta x}\partial_{\theta} v^h_{i,t}\partial_{x} v^h_{j,l}$ can be evaluated as
\begin{align}
    &\int_{\Omega^h} D_{\theta x}\partial_{\theta} v^h_{i,t}\partial_{x} v^h_{j,l}\,d\theta dxdy \nonumber\\
    &= \int_{\Omega^h}D_{\theta x}\,\phi_i(x, y)k(\theta, \xi_l)\,\partial_{\theta} k(\theta, \xi_t)\partial_{x} \phi_j(x, y)\,d\theta dxdy \nonumber\\
    &=\int_{\Omega^h_{xy}} \hat{D}_{\theta x}\,\phi_i(x, y)\,\partial_{x} \phi_j(x, y)\,dxdy, 
\end{align}
where 
\begin{align}\label{type-1-integral}
  \hat{D}_{\theta x} = \int_{\Omega^h_{\theta}}D_{\theta x}\, k(\theta, \xi_l)\partial_{\theta} k(\theta, \xi_t)\,d\theta.
\end{align}
Assume that we take Gaussian kernel as in previous integral, then
\begin{align}
   &\hat{D}_{\theta x}\nonumber\\ 
   &= (2\pi\kappa^2)^{-1}\int_{\Omega^h_{\theta}} \kappa^{-2}D_{\theta x}(\theta - \xi_t)\nonumber\\ 
   &\exp\left(-\frac{1}{2\kappa^{2}}\left(2\,( \theta - \frac{\xi_t + \xi_l}{2} )^2 + \frac{1}{2}(\xi_t - \xi_l)^2\right)\right)\,d\theta\nonumber\\
   &=\exp\left(-\frac{1}{4\kappa^{2}}(\xi_t -\xi_l)^2\right)\times(2\pi\kappa^4)^{-1}\times\nonumber\\
   &\int_{\Omega^h_{\theta}} D_{\theta x}(\theta - \xi_t) \exp\left(-\frac{1}{\kappa^{2}}\,( \theta - \frac{\xi_t + \xi_l}{2} )^2\right)\,d\theta\nonumber\\
   &\approx \frac{1}{4\sqrt{\pi}{\kappa}^{3}}\,D_{\theta x} (\xi_l - \xi_t)\exp\left(-\frac{1}{4{\kappa}^{2}}(\xi_t -\xi_l)^2\right),
\end{align}
where $D_{\theta x}\triangleq D_{\theta x}(\theta, x, y)$ should be evaluated at $\theta=\frac{\xi_t +\xi_l}{2}$ in the last equation, i.e., $D_{\theta x}(\frac{\xi_t +\xi_l}{2}, x, y)$. 

The integral of $D_{x x}\partial_{x} v^h_{i,t}\partial_{x} v^h_{j,l}$ can be evaluated as
\begin{align}
    &\int_{\Omega^h} D_{x x}\partial_{x} v^h_{i,t}\partial_{x} v^h_{j,l}\,d\theta dxdy \nonumber\\
    &= \int_{\Omega^h}D_{x x}\,k(\theta, \xi_t)k(\theta, \xi_l)\,\partial_{x}\phi_i(x, y) \partial_{x} \phi_j(x, y)\,d\theta dxdy \nonumber\\
    &=\int_{\Omega^h_{xx}} \hat{D}_{x x}\,\partial_{x} \phi_i(x, y)\,\partial_{x} \phi_j(x, y)\,dxdy, 
\end{align}
where 
\begin{align}\label{type-2-integral}
  \hat{D}_{x x} = \int_{\Omega^h_{\theta}}D_{x x}\, k(\theta, \xi_l) k(\theta, \xi_t)\,d\theta.
\end{align}
Assume that we take Gaussian kernel as in previous integral, then
\begin{align}
   &\hat{D}_{x x} \nonumber\\
   &= (2\pi\kappa^2)^{-1}\times\nonumber\\
   &\int_{\Omega^h_{\theta}} D_{x x}\exp\left(-\frac{1}{2{\kappa}^{2}}\left(2\,( \theta - \frac{\xi_t + \xi_l}{2} )^2 + \frac{1}{2}(\xi_t - \xi_l)^2\right)\right)\,d\theta\nonumber\\
   &=\exp\left(-\frac{1}{4{\kappa}^{2}}(\xi_t -\xi_l)^2\right)\times(2\pi{\kappa}^2)^{-1}\times\nonumber\\
   &\int_{\Omega^h_{\theta}} D_{x x}\exp\left(-\frac{1}{{\kappa}^{2}}\,( \theta - \frac{\xi_t + \xi_l}{2} )^2\right)\,d\theta\nonumber\\
   &\approx \frac{1}{2\sqrt{\pi}{\kappa}}\,D_{x x} \exp\left(-\frac{1}{4{\kappa}^{2}}(\xi_t -\xi_l)^2\right),
\end{align}
where $D_{\theta x}\triangleq D_{\theta x}(\theta, x, y)$ should be evaluated at $\theta=\frac{\xi_t +\xi_l}{2}$ in the last equation, i.e., $D_{\theta x}(\frac{\xi_t +\xi_l}{2}, x, y)$.

For the second term inside the integral of $K_{(i, t), (j, l)}$, we have
\begin{align}
&\int_{\Omega^h} v^h_{i,t}\,q^T\,\nabla v^h_{j, l}\,d\theta dxdy = \nonumber\\
&\int_{\Omega^h} k(\theta, \xi_t)\phi_i(x, y)
  (q_{\theta}, q_x, q_y) \left(\partial_{\theta} v^h_{j, l},  \partial_{x}v^h_{j, l}, \partial_{y}v^h_{j, l}  \right)^T \,d\theta dxdy \nonumber\\
  &= \int_{\Omega_{xy}^h} \left(\int_{\Omega_{\theta}^h} q_{\theta} k(\theta, \xi_t)\partial_{\theta} k(\theta, \xi_l)\,d\theta\right) \phi_i(x, y)\phi_j(x, y)\,dxdy\nonumber\\
  &+ \int_{\Omega_{xy}^h} \left(\int_{\Omega_{\theta}^h} q_{x} k(\theta, \xi_t)k(\theta, \xi_l)\,d\theta\right) \phi_i(x, y)\partial_{x}\phi_j(x, y)\,dxdy\nonumber\\
   &+ \int_{\Omega_{xy}^h} \left(\int_{\Omega_{\theta}^h} q_{y} k(\theta, \xi_t)k(\theta, \xi_l)\,d\theta\right) \phi_i(x, y)\partial_{y}\phi_j(x, y)\,dxdy,
\end{align}
where the kernel integrals may be approximated by the property of kernels. The approximations to three integrals (with respect to $\theta$) inside the above equation are similar to integral $(\ref{type-1-integral})$ and $(\ref{type-2-integral})$.

The final term in the stiffen matrix
\begin{align}
    &\int_{\Omega^h} b\,v^h_{i, t}\,v^h_{j, l}\,d\bm{x} \nonumber\\
    &= \int_{\Omega_{xy}^h} \left(\int_{\Omega_{\theta}^h}b\,k(\theta, \xi_t)k(\theta, \xi_l)\,d\theta\right) \phi_i(x, y)\phi_j(x, y)\,dxdy,
\end{align}
where the approximation to inside integrals (with respect to $\theta$) is similar to integral $(\ref{type-2-integral})$.

\subsubsection{Construction of Constant Vector $F$}~\\
For the constant $F$, we have
\begin{align}
    F_{i,t} &= \int_{\Omega^h} f\,v^h_{i,t}\,d\bm{x} \nonumber\\ 
    &= \int_{\Omega_{xy}^h} \left(\int_{\Omega_{\theta}^h} f(\theta, x, y)\,k(\theta, \xi_t)\,d\theta\right) \phi_i(x, y)\,dxdy \nonumber\\ 
    &{\mbox{(using Gaussian kernel)}}\nonumber\\
    &\approx \int_{\Omega_{xy}^h} f(\xi_t, x, y) \phi_i(x, y)\,dxdy
\end{align}

\subsection{Appendix to Section~\ref{tensor-product-basis}}\label{app:stiffen-matrix-elements}

For the second term in Eq.~(\ref{diffusion-term-tensor-basis}) 
in the paper
that involves $D_{qp}$, we can treat the integral similarly
\begin{align}\label{diffusion-term:second-term-integral}
    &\int_{\Omega^h} \nabla_{x^q}v^h_{i, t}\cdot D_{qp} \nabla_{x^p}v^h_{j, l}\,dx
    \\\nonumber
    &=\int_{\Omega^h_p} \hat{D}_{qp}\phi_i(x^p)\nabla_{x^p}\phi_j(x^p)\,dx^p,
\end{align}
where once again $\hat{D}_{qp}$ is approximated for Gaussian kernels as
\begin{align}
   \hat{D}_{qp}&\sim_{c,\kappa} D_{qq} (\xi_l - \xi_t) \exp\left(-\frac{1}{4\kappa^{2}}\lVert \xi_t -\xi_l\rVert^2\right).
\end{align}
One can see the computation of Eq.~(\ref{diffusion-term:second-term-integral})
is in the same way to that of Eq.~(\ref{diffusion-term:first-term-integral}) in the paper. 

For the second term inside the integral of $K_{\hat{i},\hat{j}}$ in Eq.~(\ref{tensor-stiffen-K}) in the paper 
, we have
\begin{align}
&\int_{\Omega^h} v^h_{i,t}\,q^T\,\nabla v^h_{j, l} \,d\bm{x} = \nonumber\\
  &= \int_{\Omega_p^h} (\int_{\Omega_q^h} q_q k(x^q, \xi_t)\nabla_{x^q} k(x^q, \xi_l)\,dx^q ) \phi_i(x^p)\phi_j(x^p)\,dx^p\nonumber\\
  &+ \int_{\Omega_p^h} (\int_{\Omega_q^h} q_p k(x^q, \xi_t) k(x^q, \xi_l)\,dx^q ) \phi_i(x^p)\nabla_{x^p}\phi_j(x^p)\,dx^p.
\end{align}
The final term inside the integral of $K_{\hat{i},\hat{j}}$
\begin{align}\label{mixed-stiffen-K}
    &\int_{\Omega^h} b\,v^h_{i, t}\,v^h_{j, l}\,dx \nonumber\\
    &= \int_{\Omega_p^h} \left(\int_{\Omega_q^h}b\,k(x^q, \xi_t)k(x^q, \xi_l)\,dx^q\right) \phi_i(x^p)\phi_j(x^p)\,dx^p.
\end{align}
The inner integrals in the above equations can be approximated similarly to those in Section~\ref{tensor-product-basis}. 

\subsection{Analysis of the Proposed Approach}\label{appendix:Method-Analysis}

To analyze the stability of our method, we show that the value function in the proposed approach consists of two desired properties. 
One is that the maximum of the value function is at goal and no local maximum inside the feasible motion region. The other is that the derived trajectories of the robot always leads to the goal {without navigating into unsafe regions}. The proposed hybrid representation of the value function thereby preserves these two advantages. Below we provide a heuristic argument.

\textbf{Maximum of value function at goal}. In our design, $R(x, a)$ is set to $0$ in most feasible regions and set to a positive value only around the goal state when action leads to the goal. In view of Equation~\eqref{eq:value-fn-policy}, the value function in an MDP is positive under the assumption of the positive transition probability from a state to another in the feasible region. Since the  DMDP is the continuous limit to the MDP, the optimal value function $v^{\pi}$ in Equation~\eqref{eq:strong-form-optimality} in the paper must be positive too (note that it cannot be zero). Firstly, let us consider the feasible state region $\Omega^0$ of $R(x, a)=0$ for any action. If there is a local maximum inside $\Omega^0$, then we must have $\nabla v^{\pi} = 0$ and $\nabla^2 v^{\pi} \leq 0$. Accordingly, we have $\nabla\cdot \sigma_x^\pi\nabla v^{\pi}\leq 0$~\citep[see][Chap.~6.4.1]{EvensPDE2010}. Since $v^{\pi}>0$, it is impossible for Equation~\eqref{eq:strong-form-optimality} in the paper to hold. Therefore, the optimal value function cannot attain its local maximum inside $\Omega^0$. The maximum value only occurs at the boundary of $\Omega^0$. Secondly, consider the region $\Omega^{>0}$ of $R(x, a)>0$ for an action, which also includes any boundary between the interior of $\Omega^{>0}$ and of $\Omega^{0}$. Such region is close to the goal state. As in one of boundary conditions we constrain $v^{\pi}$ to be a very large positive value at the goal, and this goal value is significantly larger than any reward. In view of the Bellman optimality equation, we can see that the maximum value of $v^{\pi}$ is at the goal state. 

\textbf{Trajectory toward the goal from a state}. One key question is whether the optimal value function can derive a series of actions that lead a robot to the goal starting from any state in the feasible region. We assume that the values of $\mu_x^\pi/\|\mu_x^\pi\|$ consists of a $d$-dimensional unit ball for all possible policies. That is, for any direction in the $d$-dimensional space there is an action such that $\mu_x^\pi$ falls in that direction. If we additionally assume that first order of value function dominates its second order, then we must have $(\mu_x^\pi)^T\nabla v^{\pi}(x) > 0$ in $\Omega^0$ for the optimal value function in Equation~\eqref{eq:strong-form-optimality} in the paper. This means that the action in $\Omega^0$ toward the increase direction of the value function. For the region $\Omega^{>0}$, because we design $R(x, a) > 0$ for actions leading to the goal, and because the goal is the maximum point of the value function, we thus also have $(\mu_x^\pi)^T\nabla v^{\pi}(x) > 0$ in Equation~\eqref{eq:strong-form-optimality} in $\Omega^{>0}$. In sum, our approach can produce a sequence of robotic actions to the goal starting from any state in the feasible region.


\vfill

\end{document}